\documentclass{article}

\PassOptionsToPackage{numbers, compress}{natbib}

\usepackage[preprint]{neurips_2026}

\usepackage[utf8]{inputenc}
\usepackage[T1]{fontenc}
\usepackage{hyperref}
\usepackage{url}
\usepackage{booktabs}
\usepackage{amsfonts}
\usepackage{nicefrac}
\usepackage{microtype}
\usepackage[table,dvipsnames]{xcolor}
\usepackage{graphicx}
\usepackage{multirow}
\usepackage{colortbl}
\usepackage{array}
\usepackage{adjustbox}
\usepackage{amssymb}
\usepackage{enumitem}
\usepackage{amsmath}
\usepackage{pifont}
\usepackage{wrapfig}
\usepackage{titlesec}

\definecolor{tableheader}{HTML}{E8EEF7}
\definecolor{propshade}{HTML}{EAF3FB}
\definecolor{opensrc}{HTML}{F0F7EC}
\definecolor{best}{HTML}{2C5AA0}
\definecolor{second}{HTML}{888888}
\definecolor{taskvideo}{HTML}{EAF2FA}
\definecolor{taskprompt}{HTML}{F4EEE0}
\definecolor{bestcell}{HTML}{2C5F7A}
\definecolor{secondcell}{HTML}{8B9DC3}
\definecolor{checkcolor}{HTML}{2C5F7A}
\definecolor{groupshade}{HTML}{EBEBEB}
\definecolor{invcol}{HTML}{DBE7F5}

\newcommand{\cmark}{\textcolor{ForestGreen!85!black}{\ding{51}}}
\newcommand{\xmark}{\textcolor{black!25}{\ding{55}}}
\newcommand{\IC}{\cellcolor{invcol}}

\titlespacing*{\paragraph}{0pt}{0.5ex}{0.5em}

\title{VI-Bench: Benchmarking Prompt Inversion from AIGC Videos}

\author{%
  Wulin Xie$^{1,2}$ \quad
  Rui Zhao$^{3}$ \quad
  Kecen Li$^{4}$ \quad
  Xiujin Liu$^{5}$ \\
  \textbf{Bokang Zhang}$^{6}$ \quad
  \textbf{Zheng Liu}$^{3}$ \quad
  \textbf{Xinwen Hou}$^{1,2}$ \quad
  \textbf{Chen Gong}$^{3}$\thanks{Corresponding author.} \\[0.5em]
  $^{1}$Institute of Automation, Chinese Academy of Sciences \\
  $^{2}$University of Chinese Academy of Sciences \\
  $^{3}$University of Virginia \\
  $^{4}$National University of Singapore \\
  $^{5}$University of Michigan, Ann Arbor \\
  $^{6}$The Chinese University of Hong Kong, Shenzhen
}

\begin{document}

\maketitle

\begin{abstract}
Recent advances in video generation have made prompt-based control increasingly central to AIGC video generation.  Prompts specify what a video should depict and how it should be represented, controlling factors such as visual style or camera behavior. 
%Generated videos may expose information about their underlying prompts.  
Understanding this recoverability is important both for creative reuse and editing, and for assessing prompt leakage risks. However, existing video understanding benchmarks do not measure this capability: a caption may describe what is visible, but a replayable prompt must recover the generation-relevant controls needed to reproduce the video. To address this gap, we introduce VI-Bench, a benchmark built from 16.1 million real-user prompts and 900 human-verified AIGC videos. VI-Bench spans three progressively harder settings, namely single-shot semantic grounding, control over style and camera behavior, and multi-shot compositional inversion, and evaluates five generation-critical dimensions: subject, action, scene, style, and camera. We evaluate 18 representative VLMs, including 2 proprietary and 16 open-source models on VI-Bench, using an Inversion Score that measures prompt-level alignment with the original prompt and video-level fidelity of the regenerated video. The results reveal substantial limitations: even the strongest model achieves only 0.632 on Inversion Score, performance degrades sharply as samples require richer control and multi-shot reasoning, and models often produce plausible prompts whose regenerated videos deviate from the reference. These findings show that video prompt inversion is a distinct and under-evaluated capability requiring models to transform visual understanding into replay-stable generative control.  
\end{abstract}

\begin{figure*}[h]
  \centering
  % \vspace{-3pt}  % 在caption后添加，缩小与正文的间距
\includegraphics[width=0.95\linewidth]{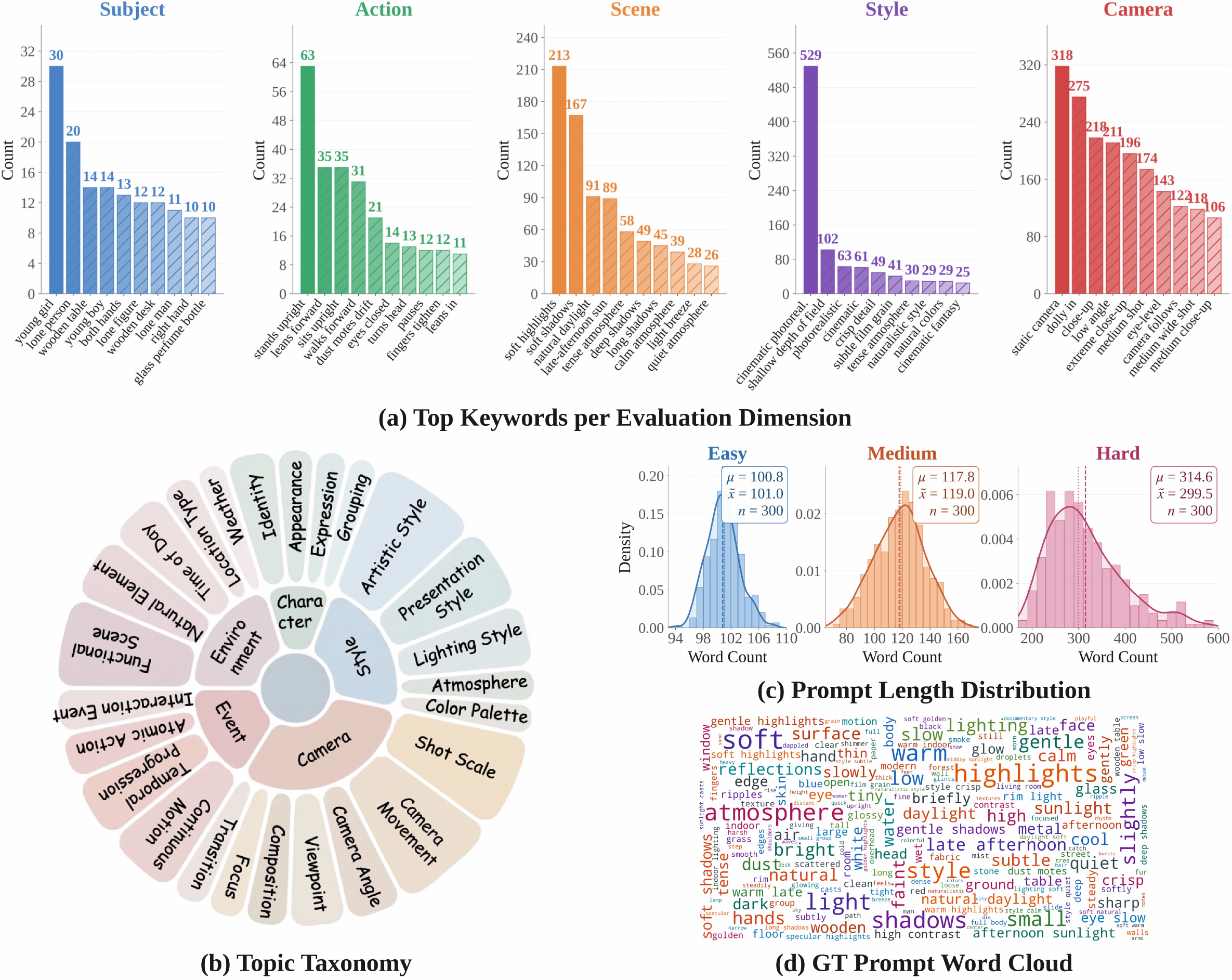}
\caption{Statistical distributions of VI-Bench.}
\label{fig:dataset_overview}
\vspace{-20pt}  % 在caption后添加，缩小与正文的间距
\end{figure*}

\section{Introduction}
Recent advances in video generation have enabled the synthesis of coherent and high-quality AIGC videos~\cite{seedance20,helios,MAGI-1,SkyReels-V2,Step-Video-T2V}, making prompt-based control increasingly central to controllable video generation. 
Prompts specify not only what a video should depict, but also how it should be presented, controlling factors such as visual style, camera behavior, and temporal composition~\cite{ma2026controllablevideogenerationsurvey,MIMO}. 
In real-world creative platforms, users may encounter a compelling AI-generated video and wish to reproduce its visual style, camera motion, or temporal structure for editing or creative reuse, while the original prompt remains unavailable. 
The emergence of prompt marketplaces such as PromptBase~\cite{PromptBase} and PromptAi Market~\cite{promptaimarket}, where users can buy and sell prompts for AIGC video generation, further shows that prompts have become valuable creative assets rather than disposable text. 
At the same time, generated videos may unintentionally reveal information about their underlying prompts, raising practical concerns about prompt leakage, proprietary prompt templates, and the protection of generation workflows. 
Understanding how recoverable such prompt-level controls are is therefore important not only for creative reuse, but also for assessing prompt leakage risks and designing defenses against prompt leakage. 
This dual role of video prompt recovery raises a fundamental question: can we measure whether a generated video exposes a replayable prompt, namely a prompt that can be executed by a video generator to reproduce the reference video?

Prior efforts have explored prompt recovery from generated content through prompt stealing, template extraction, and optimization-based inversion~\cite{ARPO,EvoStealer,PromptStealer,VGD}. 
However, these studies focus on designing specific recovery algorithms or analyzing attack cases in text-to-image settings, rather than providing a systematic benchmark for measuring how much prompt-level control information can be recovered from generated videos. 
More importantly, image-centric inversion methods do not naturally transfer to the video domain. 
Unlike images, videos require recovering generative factors that evolve over time, including temporal progression, motion continuity, stylistic consistency, camera dynamics, and multi-shot narrative structure~\cite{NOVA,MotionPro}. 
Although Vision-Language Models (VLMs)~\cite{Qwen25-VL,bai2025qwen3vltechnicalreport,InternVL3,Keye-VL} can be used for video prompt inversion in practice, this emerging use case remains poorly defined and lacks dedicated evaluation. 
This raises a key question: \textit{how can we systematically evaluate the ability of VLMs to infer replayable generation controls from AIGC videos, and where do current models fail?}

While recent VLMs have achieved strong performance on video understanding and captioning~\cite{Video-MME,Video-MME-v2,VidCapBench}, these tasks primarily assess descriptive understanding, namely whether a model can recognize, interpret, or describe what is visible in a video. 
In other words, a video caption answers what is visible to a human observer, whereas an inversion prompt must specify what a generator should execute to reproduce the video. 
Thus, video prompt inversion requires models to identify not only visible semantics, but also generation-relevant controls such as subject identity, action dynamics, scene layout, visual style, camera behavior, and temporal structure. 
This distinction also changes how the task should be evaluated. 
Because valid prompts may differ in wording while remaining equally effective for reproduction, text similarity alone cannot serve as a reliable metric~\cite{MIMO,ma2026controllablevideogenerationsurvey}. 
Conversely, two textually similar prompts may produce noticeably different videos once executed by a generator.
As a result, a proper evaluation should go beyond text comparison and test whether the inferred prompt can actually reproduce the reference video under replay.

To address this gap, we introduce \textbf{VI-Bench}, a dedicated benchmark for evaluating whether VLMs can recover replayable prompts from AIGC videos. 
VI-Bench is built from \textbf{16.1M} real-user prompts~\cite{DiffusionDB,VidProM,TIP-I2V}, which are cleaned into approximately 3.9M high-quality prompts and organized into topic pools for benchmark construction. 
The final benchmark contains 900 human-verified AIGC videos generated by two video generators and organized into three progressively harder settings: single-shot semantic grounding, control over style and camera behavior, and multi-shot compositional inversion. 
Each sample is evaluated along five generation-critical dimensions: Subject, Action, Scene, Style, and Camera. 
Beyond dataset construction, VI-Bench evaluates whether a recovered prompt is not only aligned with the original prompt, but also effective when executed by the generator to reproduce the reference video. 
In this way, VI-Bench aims to measure prompt recoverability and replay-oriented generative control, rather than descriptive video understanding alone. 
We conduct an extensive evaluation of 18 representative VLMs, including 2 proprietary and 16 open-source models on VI-Bench. 
Using the Inversion Score that jointly measures prompt-level alignment with the original prompt and video-level fidelity after replay, our benchmark analysis leads to the following findings:
\begin{itemize}[leftmargin=*]
\item \textbf{Current VLMs remain far from solving the video prompt inversion task.} Even the strongest model achieves only 0.632 overall Inversion Score on a normalized $[0,1]$ scale, showing that current models still fail to reliably recover prompts that are both faithful to the original prompt and effective under replay. Moreover, the performance of most models drops markedly as samples move from single-shot to multi-shot videos.
    \item \textbf{Video prompt inversion is not equivalent to video understanding or captioning.} Strong performance on existing video understanding benchmarks does not necessarily translate into strong video prompt inversion, suggesting that recoverability is distinct from descriptive video understanding. Moreover, models often produce prompts that appear semantically plausible, but the videos regenerated from these prompts still deviate substantially from the references.

    \item \textbf{The main failures are factor-dependent and become more noticeable in multi-shot videos.} Subject and Style exhibit the largest prompt-to-replay gaps, suggesting that models may recognize what should be recovered but fail to express it in a replay-stable form. Multi-shot videos further expose a major capability boundary, where models must aggregate information across shots while preserving temporally consistent generative controls.
\end{itemize}

\begin{figure*}[!t]
  \centering
  \includegraphics[width=\linewidth]{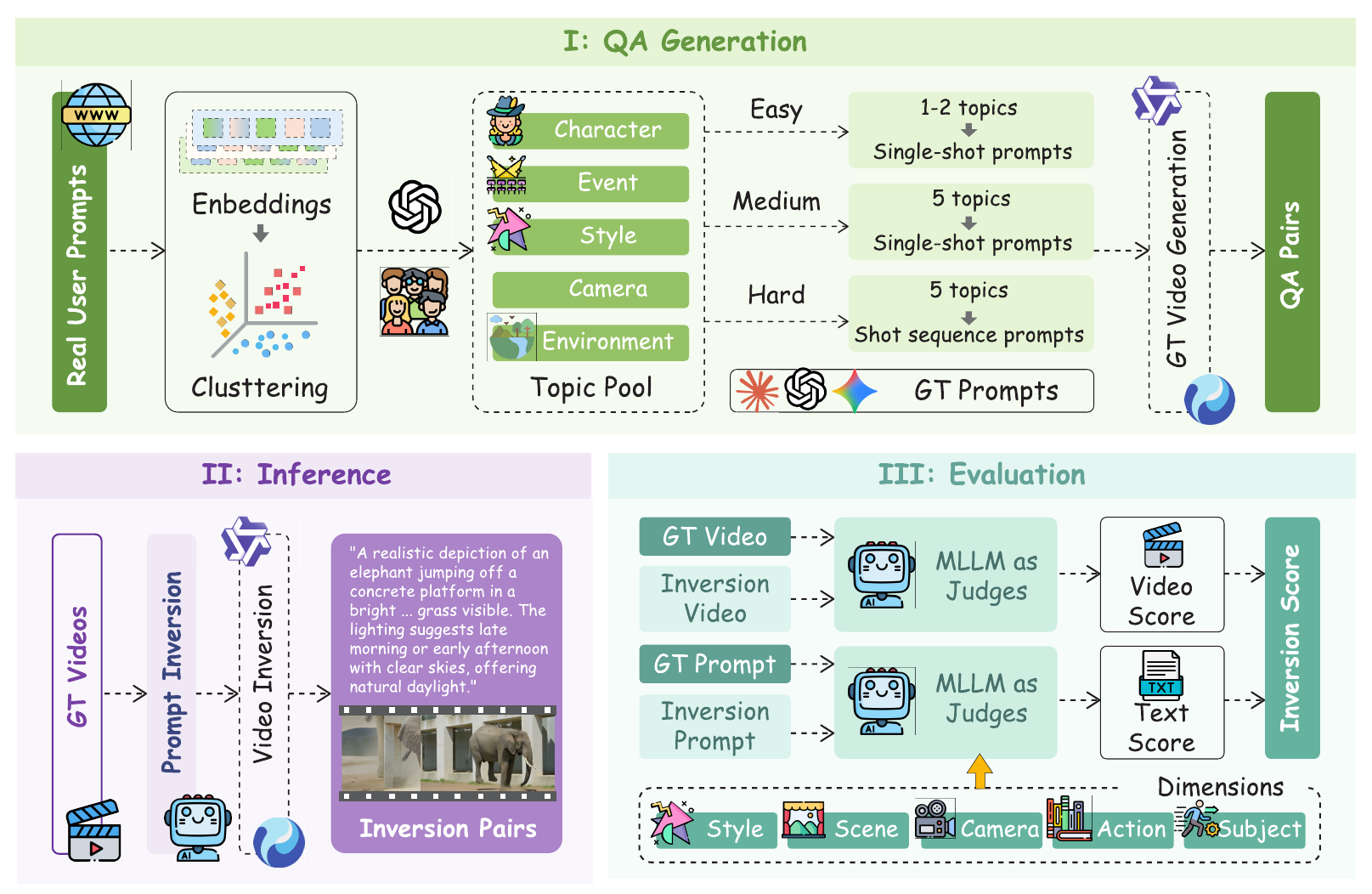}
  \caption{\textbf{Overview of the VI-Bench pipeline.} Real-user prompts are clustered into topic pools to synthesize Easy, Medium, and Hard prompts, which are rendered into ground-truth videos. VLMs recover prompts from these videos, and the recovered prompts are replayed and evaluated across five dimensions to compute the final Inversion Score.}
  \label{fig:visualization}
  \vspace{-20pt}
\end{figure*}

\section{Related Work}

\noindent \textbf{Image Prompt Inversion.}
Image prompt inversion aims to recover a text prompt from a reference image such that the prompt can reproduce similar content and style~\cite{PromptStealer,EvoStealer,ARPO,VGD}. 
Existing studies mainly focus on text-to-image generation. 
VGD~\cite{VGD} studies the recovery of readable prompts from generated images by combining language-model-based prompt generation with visual feedback from CLIP~\cite{CLIP}. 
ARPO~\cite{ARPO} formulates reverse prompt engineering as an iterative optimization process that refines prompts through repeated image generation and comparison with the reference image. Security-oriented studies further examine prompt stealing risks in text-to-image systems, with PromptStealer~\cite{PromptStealer} investigating whether key prompt components, such as subjects and modifiers, can be inferred from generated images, while EvoStealer~\cite{EvoStealer} further studies the stealing of reusable prompt templates from multiple generated images.

In contrast, this paper studies video prompt inversion. 
Rather than designing an attack against a specific image generator, we evaluate whether current VLMs can infer replayable generation controls from AIGC videos. 
This setting is substantially more challenging than image prompt inversion, since videos require recovering temporally structured factors such as motion continuity, camera dynamics, style consistency, and multi-shot composition.

\noindent \textbf{Benchmarking Vision-Language Models.}
Recent benchmarks evaluate Vision-Language Models from a broad range of perspectives, including general multimodal perception and reasoning~\cite{MMBench,MMVU}, video understanding and temporal reasoning~\cite{Video-MME,Video-MME-v2,MVBench,MLVU,LongVideoBench}, and video captioning for controllable text-to-video generation~\cite{VidCapBench}. 
These benchmarks have substantially advanced the evaluation of VLMs by measuring whether models can recognize visual content, reason about temporal events, answer video questions, or produce descriptive captions. However, existing benchmarks mainly evaluate descriptive video understanding, i.e., whether models can answer questions or describe visible content. 
In contrast, VI-Bench evaluates whether a model can recover a replayable prompt from an AIGC video. 
The recovered prompt is executed by the original generator, and evaluation jointly measures prompt fidelity and replay fidelity across Subject, Action, Scene, Style, and Camera.
\section{VI-Bench}

\subsection{Task Formulation}

We formulate video prompt inversion as the task of recovering a replayable prompt from a reference AIGC video. Unlike video captioning, the output is not a free-form description of visible content, but a generator-ready prompt that should preserve the controls needed for regeneration. Each sample consists of a ground-truth video $V_{\mathrm{gt}}$, its original prompt $P_{\mathrm{gt}}$, and the corresponding video generator $G$. Given only $V_{\mathrm{gt}}$, an inversion model $M$ predicts an inversion prompt, $P_{\mathrm{inv}} = M(V_{\mathrm{gt}}).$ The predicted prompt is then fed back to the original generator to produce an inversion video, $V_{\mathrm{inv}} = G(P_{\mathrm{inv}}).$ This replay step is central to our formulation: it tests whether the recovered prompt is not only semantically plausible, but also executable by the generator. We evaluate inversion quality from two perspectives: (1) prompt fidelity, measured by the similarity between $P_{\mathrm{inv}}$ and $P_{\mathrm{gt}}$, and 
(2) replay fidelity, measured by the similarity between $V_{\mathrm{inv}}$ and $V_{\mathrm{gt}}$. Both perspectives are necessary. Prompt fidelity checks whether the model recovers the original generation intent and prompt-level controls, while replay fidelity checks whether these controls actually work when executed by the generator. Using replay fidelity alone may overestimate inversion quality, since generator priors or randomness can produce a visually similar video even when the inferred prompt misses key original controls. Using prompt fidelity alone is also insufficient, since a semantically plausible prompt may still fail to reproduce the reference video.

\subsection{Data Construction Process}
We design a systematic data construction pipeline, as illustrated in Figure~\ref{fig:visualization}. The goal of this pipeline is to preserve the diversity of real user generation intents while enabling controlled construction over topic composition, difficulty level, and generator source. The process consists of four key stages: (1) \textit{Prompt Collection \& Cleaning}, (2) \textit{Topic Pool Construction}, (3) \textit{Difficulty-aware Prompt and Video Synthesis}, and (4) \textit{Human Verification}. Prompt collection anchors the benchmark in real-world prompt distributions; topic construction organizes the cleaned prompts into topic pools for sampling; difficulty-aware synthesis combines these factors into progressively harder videos; and human verification ensures that the final videos faithfully reflect their prompts.

\noindent \textbf{Prompt Collection \& Cleaning.}
We collect real-user prompts from three public datasets: DiffusionDB~\cite{DiffusionDB} (6M), VidProM~\cite{VidProM} (5M), and TIP-I2V~\cite{TIP-I2V} (5.1M), yielding 16.1M raw prompts in total. These datasets cover a broad range of public user generation scenarios, including text-to-image, text-to-video, and image-to-video prompting, providing a diverse source of real generation intents. We then filter platform prefixes, generation parameters, negative prompts, noisy expressions, non-English content, and NSFW prompts, while preserving the underlying generation intent. This process yields approximately 3.9M cleaned prompts, corresponding to a retention rate of 24.2\%, which are used for subsequent topic construction.

\noindent \textbf{Topic Pool Construction.}
To obtain diverse prompts for benchmark construction, we avoid directly sampling prompts at random, since random sampling would make it difficult to balance semantic coverage and difficulty level. Instead, we organize the cleaned prompts into topic pools. Specifically, we first encode the cleaned prompts using Qwen3-Embedding-4B~\cite{Qwen3Embedding} and then perform topic discovery with BERTopic~\cite{BERTopic}. We next use GPT-4o~\cite{GPT-4o} to assign each discovered topic to one of six categories: \textit{Character}, \textit{Event}, \textit{Style}, \textit{Environment}, \textit{Camera}, or \textit{Untagged}. These categories are chosen to reflect common prompt-level controls in video generation: Character, Event, and Environment describe core semantic content, while Style and Camera capture generator-sensitive appearance and cinematographic controls. The Untagged category is used to filter topics that are ambiguous, overly noisy, or unsuitable for controlled benchmark construction. Only the first five categories are retained as benchmark topic pools. Finally, we apply CLIP-based deduplication with a cosine similarity threshold of 0.9 to remove semantically overlapping topics, retaining the shorter topic name when duplicates are detected. This process generates approximately 800 topics in each pool, which serve as the basis for subsequent difficulty-aware prompt synthesis.

\paragraph{Difficulty-Aware Prompt and Video Synthesis.}
\begin{wraptable}{r}{0.42\linewidth}
\vspace{-12pt}
\centering
\caption{\textbf{Difficulty design of VI-Bench.} Each level progressively introduces additional generation-critical factors.}
\label{tab:difficulty_design}
\vspace{2pt}
\resizebox{\linewidth}{!}{%
\renewcommand{\arraystretch}{1.2}
\setlength{\tabcolsep}{6pt}
\begin{tabular}{lccc}
\toprule
\rowcolor{gray!8}
\textbf{Factor} & \textbf{Easy} & \textbf{Medium} & \textbf{Hard} \\
\midrule
Subject     & \cmark & \cmark & \cmark \\
Action      & \cmark & \cmark & \cmark \\
Scene       & \cmark & \cmark & \cmark \\
Style       & \xmark & \cmark & \cmark \\
Camera      & \xmark & \cmark & \cmark \\
Multi-shot  & \xmark & \xmark & \cmark \\
\bottomrule
\end{tabular}%
}
\vspace{-10pt}
\end{wraptable}
Based on the topic pools, we synthesize benchmark samples with controlled difficulty and generator diversity. To reduce stylistic bias from any single language model, we uniformly sample from GPT-4o~\cite{GPT-4o}, Claude Sonnet 4.5~\cite{Claude45}, and Gemini 2.5 Flash~\cite{Gemini25-Flash} to generate ground-truth prompts conditioned on sampled topics. We organize VI-Bench into three difficulty levels that place progressively stronger demands on prompt inversion. The difficulty design follows the intuition that video prompt inversion becomes harder as the prompt contains more generation-critical factors and longer temporal dependencies. We summarize the three levels in Table~\ref{tab:difficulty_design}. 
Easy samples are built from \textit{Character}, \textit{Event}, and \textit{Environment} topics, primarily testing semantic grounding. Medium samples additionally introduce \textit{Style} and \textit{Camera} topics, requiring recovery of generator-sensitive control factors beyond visible content description. This level explicitly separates prompt inversion from captioning: a caption may correctly describe the subject, action, and scene, but still omit style words, shot scale, viewpoint, or camera motion that are essential for reproducing the video. Hard samples further extend Medium to coherent multi-shot narratives, where multiple shot-level prompts must be recovered under temporal continuity and cross-shot compositional constraints. We generate the resulting videos using two video generators, Wan2.2~\cite{Wan} and HunyuanVideo 1.5~\cite{HunyuanVideo15}, under fixed settings. Using two generators reduces the dependence of VI-Bench on a single generation pipeline. Each prompt produces one video, while Hard samples are generated shot by shot and concatenated into a final multi-shot sequence.

\noindent \textbf{Human Verification.}
This verification step is necessary because even high-quality video generators may omit prompt factors. Each sample is assessed independently by two annotators. Annotators check whether the generated video matches the ground-truth prompt and whether any salient subject, action, scene, style, or camera factor is missing or inconsistent. Samples marked as misaligned are regenerated by varying the seed or revising the prompt, and are then re-evaluated under the same protocol. After verification, we retain 900 benchmark samples in total, with 300 samples at each difficulty level. Easy and Medium consist of single-shot videos, while Hard contains multi-shot videos with 2, 3, or 4 shots. For Hard samples, each shot is generated from a shot-level prompt, and the generated shots are concatenated in temporal order to form one final multi-shot video.

\subsection{Evaluation}
\begin{wrapfigure}{r}{0.42\textwidth}
  \centering
  \vspace{-10pt}
  \includegraphics[width=0.40\textwidth]{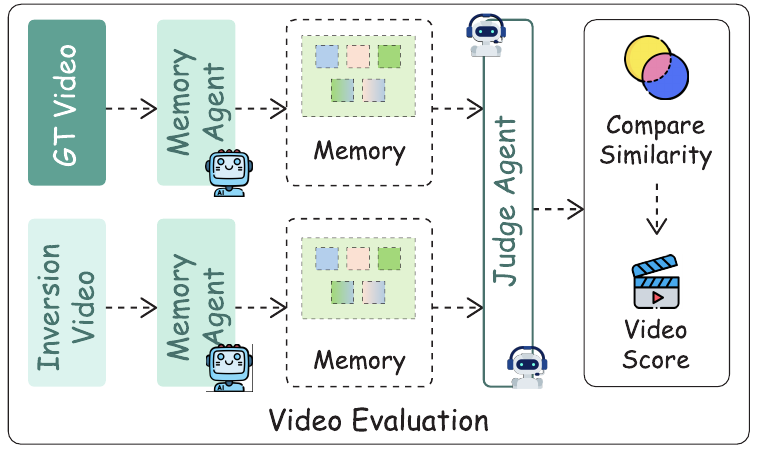}
  \caption{An overview of the video score evaluation pipeline.}
  \label{fig:video_evaluation}
  \vspace{-10pt}
\end{wrapfigure}

We evaluate video prompt inversion from two complementary perspectives: \textit{prompt fidelity} and \textit{replay fidelity}. 
This protocol is designed to measure reverse-prompting ability rather than general video understanding: the output is judged by whether it preserves the original generation intent and can be replayed by the generator. 
Given a reference video $V_{\mathrm{gt}}$, a VLM predicts a prompt $P_{\mathrm{inv}}$, which is replayed by the same generator under the original generation settings to generate an inversion video $V_{\mathrm{inv}}$. 
For Hard-level samples, $P_{\mathrm{inv}}$ contains multiple shot-level prompts. Each shot-level prompt is replayed separately using the original generator and settings, and the generated shots are concatenated in temporal order to form the full inversion video $V_{\mathrm{inv}}$. 
VI-Bench evaluates two aspects of inversion quality: whether the inferred prompt is semantically faithful to the original generation intent, and whether it remains effective for reproducing the reference video when executed by the generator.

\paragraph{Prompt-level evaluation.} To assess prompt fidelity, we compare the inferred prompt $P_{\mathrm{inv}}$ with the ground-truth prompt $P_{\mathrm{gt}}$ using GPT-4o as the judge. The two prompts are evaluated on five dimensions: $D=\{\textit{Subject}, \textit{Action}, \textit{Scene}, \textit{Style}, \textit{Camera}\}$, and \textit{Camera}. For each dimension, the judge assigns a raw score $\hat{s}^{p}_{d}\in[1,5]$, which is normalized to $s^{p}_{d}=(\hat{s}^{p}_{d}-1)/4$. 
The final Prompt Score is computed as:
\begin{equation}
S_{\mathrm{prompt}} = \frac{1}{|D|} \sum_{d \in D} s^{p}_{d},
\end{equation} where $s^{p}_{d}$ denotes the prompt-level score on dimension $d$.

\paragraph{Video-level evaluation.}
To assess replay fidelity, we compare the replay video $V_{\mathrm{inv}}$ with the reference video $V_{\mathrm{gt}}$ using a two-agent evaluation framework. 
As shown in Figure~\ref{fig:video_evaluation}, a Memory Agent first reads each video independently and summarizes it along the same difficulty-specific dimensions $D$. 
A Judge Agent then compares the two video memories and assigns a raw score $\hat{s}^{v}_{d}\in[1,5]$ for each dimension, which is normalized to $s^{v}_{d}=(\hat{s}^{v}_{d}-1)/4$. 
The final Video Score is computed as:
\begin{equation}
S_{\mathrm{video}} = \frac{1}{|D|} \sum_{d \in D} s^{v}_{d},
\end{equation}
where $s^{v}_{d}$ denotes the normalized video-level score on dimension $d$.

Finally, we define the overall Inversion Score as $S_{\mathrm{inv}} = \frac{1}{2}(S_{\mathrm{prompt}} + S_{\mathrm{video}})$, which measures whether a model can recover prompts that are both faithful and replay-effective.
\section{Experiment}
% We organize our experiments to progressively answer four questions. First, we evaluate the overall capability of current VLMs on VI-Bench across different difficulty levels. Second, we examine whether video prompt inversion can be explained by conventional video understanding or captioning ability. Third, we analyze failures across generation-critical dimensions to identify where descriptive recovery breaks down into replay failure. Finally, we study multi-shot effects to analyze how temporal complexity influences the difficulty of video prompt inversion.

\subsection{Experimental Setup}
We evaluate 18 representative VLMs on VI-Bench, including 2 proprietary VLMs, Doubao-Seed-2.0-pro~\cite{Seed20-Pro} and GPT-4o~\cite{GPT-4o}, and 16 open-source VLMs, including OmniVinci~\cite{OmniVinci}, Qwen2.5-VL series~\cite{Qwen25-VL}, Qwen3-VL series~\cite{bai2025qwen3vltechnicalreport}, Qwen3.5~\cite{qwen3.5}, VideoLLaMA3~\cite{Video-LLaMA}, Keye-VL~\cite{Keye-VL}, InternVL2.5 \& 3~\cite{InternVL25,InternVL3}, and two agent-style models, PyVision-Video~\cite{PyVision-RL} and LongVideoAgent~\cite{LongVideoAgent}. In terms of model scale, the evaluated open-source models span small (3B/4B), medium (7B/8B/9B), and large (30B/32B/72B) regimes. 

To ensure consistent comparison, all models are evaluated under a shared evaluation protocol: given a reference video, each model is prompted with the same inversion instruction and asked to generate an inversion prompt. For video input, most models follow their default or recommended configurations, while GPT-4o is limited to 50 input frames due to API constraints. For replay, the inferred prompt is fed back to the same generator used to generate the ground-truth sample, under the original generation settings and a fixed seed, to generate the replay video for evaluation. For video-level evaluation, we instantiate the Memory Agent with Qwen3-VL-8B and the Judge Agent with Qwen3.5-9B. These settings ensure that performance differences mainly reflect the models' inversion ability rather than variations in prompting or replay settings.

\subsection{Main Results}

\noindent \textbf{Experiment Design.}
We evaluate 18 proprietary and open-source VLMs on VI-Bench and report their Prompt Score, Video Score, and Inversion Score across the Easy, Medium, Hard, and Overall settings in Table~\ref{tab:main_results}. The goal is to measure whether current VLMs can recover prompts that are semantically aligned with the original prompts and are effective when replayed by the video generator.

\noindent \textbf{Result Analysis.}
Table~\ref{tab:main_results} shows that current models remain far from solving video prompt inversion: even the strongest model achieves only 0.632 overall Inversion Score, indicating that recovering replayable prompts from AIGC videos remains challenging for existing VLMs. We also observe a clear performance drop as task difficulty increases. Across nearly all models, scores decrease from Easy to Hard, showing that video prompt inversion becomes harder as control factors become richer. For example, GPT-4o drops from 0.731 on Easy to 0.441 on Hard, while Doubao-Seed-2.0-pro drops from 0.751 to 0.529. Moreover, for most models, Prompt Score is higher than Video Score, and this gap becomes larger on Medium and Hard samples. This reveals a central distinction between descriptive recovery and generative recoverability: a model may produce a prompt that looks semantically plausible, yet still fail to infer a prompt that the generator can execute faithfully.

% \documentclass[11pt]{article}
% \usepackage[paperwidth=29cm,paperheight=17cm,margin=0.6cm]{geometry}
% \usepackage{booktabs}
% \usepackage{multirow}
% \usepackage[table,dvipsnames]{xcolor}
% \usepackage{colortbl}

% % --- color palette ---
% \definecolor{tableheader}{HTML}{E8EEF7}
% \definecolor{groupshade}{HTML}{EBEBEB}
% \definecolor{invcol}{HTML}{DBE7F5}

% \newcommand{\IC}{\cellcolor{invcol}}

% \begin{document}
% \thispagestyle{empty}

\begin{table*}[t]
\centering
\caption{\textbf{Main results on VI-Bench.} We evaluate 18 VLMs on the VI-Bench across three difficulty levels.
\textit{Prompt Score} measures prompt fidelity; \textit{Video Score} measures replay fidelity;
\textit{Inversion Score} is the average of the two, reflecting a model's overall ability to recover replayable prompts from AIGC videos. \textbf{Bold} indicates the best result in each column; \underline{underline} indicates the second-best.}
\label{tab:main_results}
\resizebox{\textwidth}{!}{%
\renewcommand{\arraystretch}{1.08}
\setlength{\tabcolsep}{5pt}
\begin{tabular}{l l c ccc ccc ccc ccc}
\toprule
\multirow{2}{*}{\textbf{Method}} & \multirow{2}{*}{\textbf{Release}} & \multirow{2}{*}{\textbf{\#Params}}
& \multicolumn{3}{c}{\cellcolor{tableheader}\textbf{Easy}}
& \multicolumn{3}{c}{\cellcolor{tableheader}\textbf{Medium}}
& \multicolumn{3}{c}{\cellcolor{tableheader}\textbf{Hard}}
& \multicolumn{3}{c}{\cellcolor{tableheader}\textbf{Overall}} \\
\cmidrule(lr){4-6} \cmidrule(lr){7-9} \cmidrule(lr){10-12} \cmidrule(lr){13-15}
& & & \textbf{Prompt} & \textbf{Video} & \cellcolor{invcol}\textbf{Inv.}
    & \textbf{Prompt} & \textbf{Video} & \cellcolor{invcol}\textbf{Inv.}
    & \textbf{Prompt} & \textbf{Video} & \cellcolor{invcol}\textbf{Inv.}
    & \textbf{Prompt} & \textbf{Video} & \cellcolor{invcol}\textbf{Inv.} \\
\midrule

\rowcolor{groupshade} \multicolumn{15}{l}{\textit{\textbf{Proprietary MLLMs}}} \\

Doubao-Seed-2.0-pro   & 2025-10 & --    & \textbf{0.747} & \textbf{0.755} & \IC\textbf{0.751} & \textbf{0.649} & \textbf{0.584} & \IC\textbf{0.617} & \textbf{0.627} & \textbf{0.432} & \IC\textbf{0.529} & \textbf{0.674} & \textbf{0.591} & \IC\textbf{0.632} \\
GPT-4o                & 2024-08 & --    & \underline{0.742} & 0.721 & \IC 0.731 & \underline{0.611} & 0.528 & \IC 0.570 & 0.551 & \underline{0.331} & \IC 0.441 & \underline{0.635} & \underline{0.527} & \IC\underline{0.581} \\

\midrule

\rowcolor{groupshade} \multicolumn{15}{l}{\textit{\textbf{Open-source MLLMs}}} \\

OmniVinci             & 2025-10 & 7B    & 0.731 & 0.713 & \IC 0.722 & 0.587 & 0.525 & \IC 0.556 & \underline{0.582} & 0.318 & \IC\underline{0.450} & 0.633 & 0.519 & \IC 0.576 \\
Qwen2.5-VL-72B        & 2025-01 & 72B   & 0.721 & 0.726 & \IC 0.723 & 0.581 & 0.533 & \IC 0.557 & 0.539 & 0.313 & \IC 0.426 & 0.614 & 0.524 & \IC 0.569 \\
Qwen3-VL-8B           & 2025-09 & 8B    & 0.720 & \underline{0.746} & \IC\underline{0.733} & 0.579 & \underline{0.564} & \IC\underline{0.571} & 0.477 & 0.255 & \IC 0.366 & 0.592 & 0.522 & \IC 0.557 \\
Qwen3-VL-30B          & 2025-09 & 30B   & 0.712 & 0.729 & \IC 0.720 & 0.579 & 0.549 & \IC 0.564 & 0.472 & 0.258 & \IC 0.365 & 0.588 & 0.512 & \IC 0.550 \\
LLaVA-Video           & 2024-10 & 7B    & 0.722 & 0.702 & \IC 0.712 & 0.561 & 0.484 & \IC 0.523 & 0.524 & 0.295 & \IC 0.409 & 0.602 & 0.494 & \IC 0.548 \\
Qwen3-VL-4B           & 2025-09 & 4B    & 0.715 & 0.720 & \IC 0.718 & 0.557 & 0.553 & \IC 0.555 & 0.468 & 0.250 & \IC 0.359 & 0.580 & 0.508 & \IC 0.544 \\
Qwen2.5-VL-32B        & 2025-03 & 32B   & 0.718 & 0.711 & \IC 0.715 & 0.577 & 0.536 & \IC 0.556 & 0.453 & 0.251 & \IC 0.352 & 0.583 & 0.499 & \IC 0.541 \\
Qwen2.5-VL-7B         & 2025-01 & 7B    & 0.717 & 0.676 & \IC 0.697 & 0.525 & 0.470 & \IC 0.497 & 0.500 & 0.259 & \IC 0.379 & 0.581 & 0.468 & \IC 0.524 \\
Qwen3.5               & 2025-11 & 9B    & 0.720 & 0.706 & \IC 0.713 & 0.595 & 0.503 & \IC 0.549 & 0.408 & 0.184 & \IC 0.296 & 0.574 & 0.464 & \IC 0.519 \\
VideoLLaMA3           & 2025-01 & 7B    & 0.686 & 0.655 & \IC 0.670 & 0.480 & 0.443 & \IC 0.462 & 0.483 & 0.241 & \IC 0.362 & 0.550 & 0.446 & \IC 0.498 \\
Keye-VL               & 2025-10 & 8B    & 0.712 & 0.697 & \IC 0.704 & 0.568 & 0.490 & \IC 0.529 & 0.331 & 0.102 & \IC 0.216 & 0.537 & 0.429 & \IC 0.483 \\
Qwen2.5-VL-3B         & 2025-01 & 3B    & 0.656 & 0.648 & \IC 0.652 & 0.478 & 0.403 & \IC 0.441 & 0.396 & 0.174 & \IC 0.285 & 0.510 & 0.408 & \IC 0.459 \\
InternVL3             & 2025-04 & 8B    & 0.392 & 0.707 & \IC 0.549 & 0.290 & 0.487 & \IC 0.388 & 0.253 & 0.220 & \IC 0.236 & 0.311 & 0.471 & \IC 0.391 \\
InternVL2.5           & 2024-12 & 8B    & 0.383 & 0.703 & \IC 0.543 & 0.285 & 0.496 & \IC 0.390 & 0.213 & 0.171 & \IC 0.192 & 0.294 & 0.457 & \IC 0.375 \\
PyVision-Video        & 2025-08 & 7B    & 0.656 & 0.471 & \IC 0.563 & 0.462 & 0.201 & \IC 0.332 & 0.298 & 0.205 & \IC 0.252 & 0.472 & 0.242 & \IC 0.357 \\
LongVideoAgent        & 2025-05 & 7B    & 0.539 & 0.571 & \IC 0.555 & 0.385 & 0.301 & \IC 0.343 & 0.168 & 0.097 & \IC 0.133 & 0.364 & 0.323 & \IC 0.343 \\

\bottomrule
\end{tabular}%
}
\vspace{-12pt}  % 在caption后添加，缩小与正文的间距
\end{table*}

% \end{document}

\subsection{Relationship Between Video Understanding and Video Prompt Inversion}

\begin{figure*}[h]
  \centering
  % \vspace{-3pt}  % 在caption后添加，缩小与正文的间距
\includegraphics[width=0.85\linewidth]{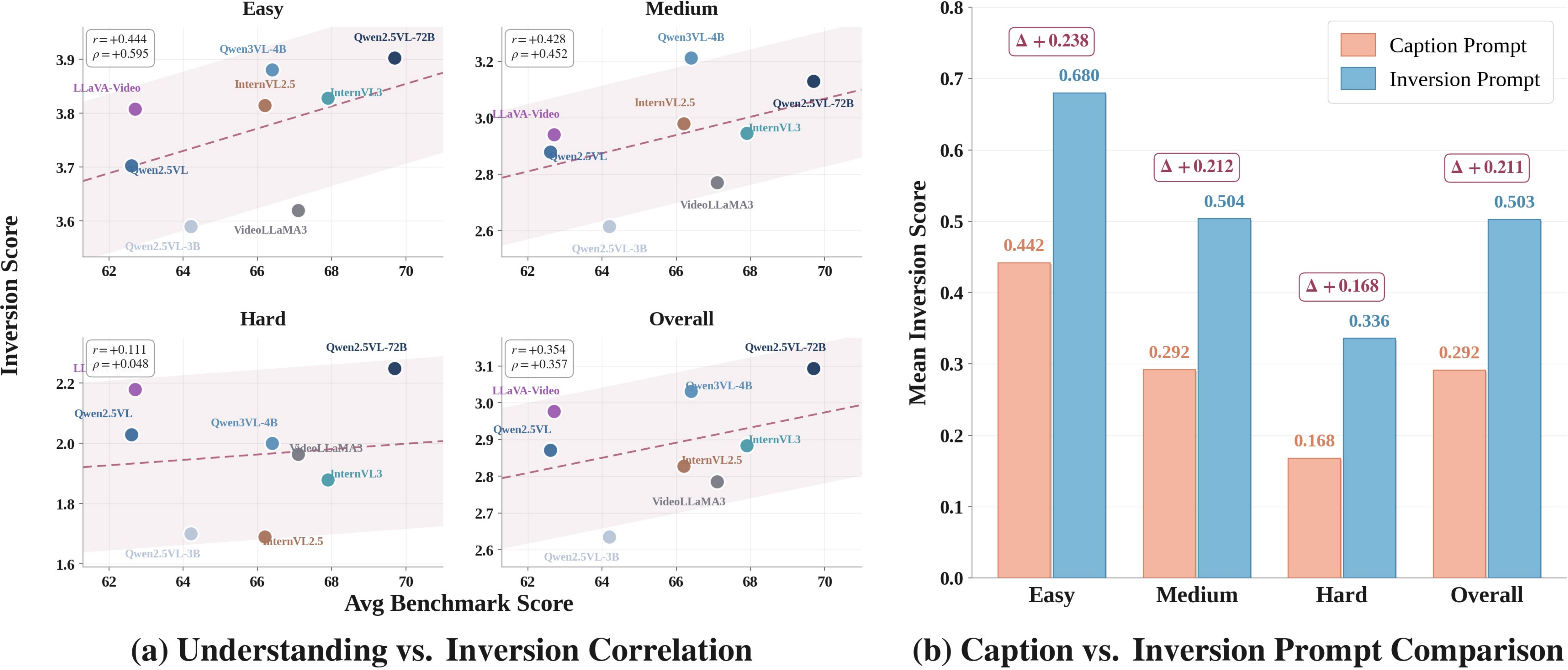}
\caption{Analysis of video understanding and the video prompt inversion task. Correlation between average scores on mainstream video understanding benchmarks and VI-Bench Inversion Scores (left), and comparison between captioning-based and inversion-based prompting (right).}
\label{fig:und_inv-analyse}
\vspace{-6pt}  % 在caption后添加，缩小与正文的间距
\end{figure*}

\noindent \textbf{Experiment Design.}
To examine whether video prompt inversion can be explained by video understanding or captioning ability, we conduct two complementary analyses, as shown in Figure~\ref{fig:und_inv-analyse}. 
First, we compare VI-Bench with existing video understanding benchmarks. 
For each model, we average its reported scores on four mainstream video understanding benchmarks, including Video-MME, MVBench, MLVU, and LongVideoBench, and correlate this average score with its VI-Bench Inversion Score under different difficulty levels. 
Each point in Figure~\ref{fig:und_inv-analyse} (left) corresponds to one model, and we report both Pearson correlation $r$ and Spearman rank correlation $\rho$. 
Second, we compare video captioning with prompt inversion. 
For the same samples, we query the same model with either a captioning system prompt or an inversion system prompt, replay both outputs using the same video generator, and evaluate the resulting videos under VI-Bench, as shown in Figure~\ref{fig:und_inv-analyse} (right).

\noindent \textbf{Result Analysis.}
\textbf{Video understanding is related to video prompt inversion, but it is not sufficient.} 
As shown in Figure~\ref{fig:und_inv-analyse} (left), general video understanding scores show only moderate correlation with VI-Bench on Easy and Medium samples, and the correlation nearly disappears on Hard samples. 
This suggests that video understanding helps models recognize what appears in the video, which explains why it still correlates with VI-Bench when the task mainly involves single-shot semantic grounding. 
However, when samples require style recovery, camera behavior, and multi-shot structure, conventional video understanding scores can no longer reliably explain inversion performance.

\textbf{Captioning is not equivalent to prompt inversion.} 
Figure~\ref{fig:und_inv-analyse} (right) shows that replacing the captioning prompt with the inversion prompt consistently improves the mean Inversion Score, from $0.442$ to $0.680$ on Easy, from $0.292$ to $0.504$ on Medium, from $0.168$ to $0.336$ on Hard, and from $0.292$ to $0.503$ overall. 
This indicates that a detailed caption may describe the video content, but it may still miss the control words needed to regenerate the video, such as style, camera motion, and temporal composition.

\subsection{Dimension-wise Analysis: From Descriptive Recovery to Generative Recoverability}
\label{sec:dimension_analysis}

\begin{figure*}[h]
  \centering
   \vspace{-10pt}  % 在caption后添加，缩小与正文的间距
\includegraphics[width=1.0\linewidth]{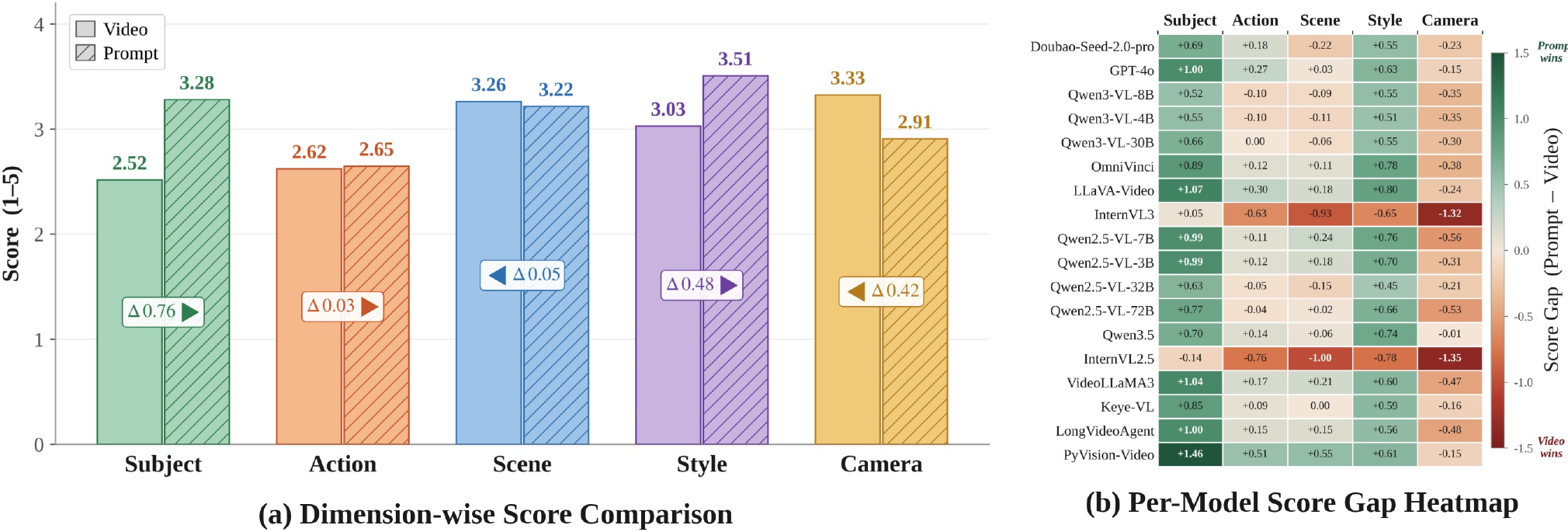}
\caption{Dimension-wise gaps between descriptive recovery and generative recoverability. Average Prompt and Video scores across the five generative dimensions (left) and per-model Prompt--Video score gaps for each dimension (right).}
\label{fig:dimension-analyse}
\vspace{-2pt}  % 在caption后添加，缩小与正文的间距
\end{figure*}

\noindent \textbf{Experiment Design.}
To better understand where current VLMs fail on video prompt inversion, we conduct two dimension-wise analyses. First, we compare the average Prompt and Video scores across the five dimensions, as shown in Figure~\ref{fig:dimension-analyse}(a). Second, we examine the per-model Prompt--Video gap for each dimension, as shown in Figure~\ref{fig:dimension-analyse}(b), to determine whether failures mainly come from weak prompt recovery or from replay instability.

\noindent \textbf{Result Analysis.}
Figure~\ref{fig:dimension-analyse}(a) shows that Subject and Style are recovered more strongly at the prompt level than at the video level, whereas Camera shows the opposite pattern. This indicates that different factors fail in different ways. For Subject and Style, models may recover a plausible textual description, but the prompt is often not precise enough for the generator to reproduce the same subject identity or atmosphere. Camera behaves differently because models often under-specify shot scale, viewpoint, or camera motion in the prompt, while the generator may still introduce plausible camera behavior through its default priors. Figure~\ref{fig:dimension-analyse}(b) further shows that this mismatch is strongly factor-dependent: most models exhibit positive gaps on Subject, suggesting replay failure, whereas Camera shows negative gaps, suggesting that camera-related content is often under-specified in the inverted prompts. A more detailed model-level analysis is provided in Appendix~\ref{app:model_level_dimension}.

\begin{figure*}[h]
  \centering
  % \vspace{-3pt}  % 在caption后添加，缩小与正文的间距
\includegraphics[width=1.0\linewidth]{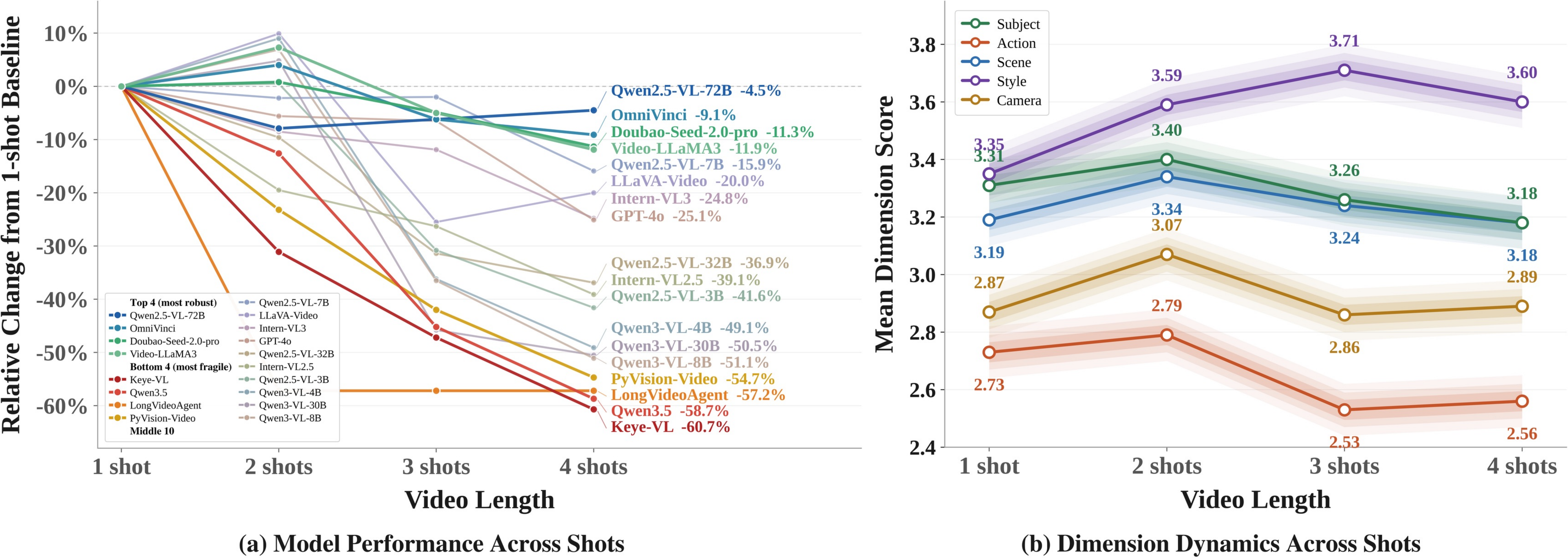}
\caption{Multi-shot effects on inversion performance. Relative performance change of each model with increasing shot number (left) and mean score trends of the five dimensions across shots (right).}
\label{fig:shot_analysis}
\vspace{-4pt}  % 在caption后添加，缩小与正文的间距
\end{figure*}

\subsection{Multi-shot Videos Reveal Capability Boundaries}

\noindent \textbf{Experiment Design.}
Long-form or film-level AIGC videos are typically composed of multiple shots rather than a single continuous scene. Therefore, prompt inversion for such videos requires more than describing one shot: the model must identify different shot segments, understand their relations, and organize them into a coherent prompt that preserves cross-shot composition and temporal continuity. To examine VLMs' prompt inversion ability on multi-shot AIGC videos, we analyze model performance across videos with different numbers of shots, as shown in Figure~\ref{fig:shot_analysis}. Figure~\ref{fig:shot_analysis}(a) reports the relative change of each model with respect to its 1-shot performance, while Figure~\ref{fig:shot_analysis}(b) shows how the mean dimension scores vary across these settings.

\noindent \textbf{Result Analysis.}
Figure~\ref{fig:shot_analysis}(a) suggests that moving to multi-shot settings does not immediately make inversion harder: several models improve from 1 shot to 2 shots, suggesting that an extra shot can provide useful visual evidence for prompt recovery. However, this benefit does not persist. From 2 shots onward, most models begin to decline, suggesting that an important difficulty arises when models must organize multiple shots into a single coherent and replayable control representation. Models also respond differently to increasing shot numbers: some remain robust or benefit from additional shots, while others degrade sharply, suggesting different abilities to exploit temporal context. Figure~\ref{fig:shot_analysis}(b) further shows that different generative factors respond differently to temporal extension. \textit{Style} consistently improves as the number of shots increases, rising from 3.35 at 1 shot to 3.71 at 3 shots, while \textit{Action} drops from 2.79 at 2 shots to 2.53 at 3 shots and remains low thereafter. \textit{Subject} and \textit{Scene} also decline after 2 shots, whereas \textit{Camera} peaks at 2 shots and then stabilizes at a lower level. These findings suggest that multi-shot inversion can be more difficult than one-shot inversion, since models must recover not only each shot, but also how the shots connect and how the whole video should be regenerated.

\section{Conclusion}
We introduce video prompt inversion as a distinct capability beyond conventional video understanding, and present VI-Bench as a dedicated benchmark for evaluating whether VLMs can recover replayable prompts from AIGC videos. Our results reveal a substantial gap between descriptive recovery and replay-stable generative control, especially under richer factors and multi-shot settings. Beyond measuring model capability, VI-Bench also offers a way to study prompt recoverability and potential prompt leakage risks in AIGC video systems. We hope VI-Bench provides a useful benchmark for future research on video prompt inversion and broader evaluation of generative understanding.

\clearpage
\newpage

\bibliography{reference}
\bibliographystyle{unsrtnat}

\clearpage
\newpage

\appendix
\begin{center}
{\LARGE \textbf{{------------Appendix------------}}}
\end{center}

\vspace{1.5em}
\noindent\textbf{\large Contents}
\vspace{1em}

{
\hypersetup{linkcolor=black}

% Section A
\noindent\makebox[1.5em][l]{\textbf{A}}\textbf{Appendix Analysis}\dotfill\pageref{app:Appendix_Analysis}

\vspace{0.8em}
% Section A
\noindent\makebox[1.5em][l]{\textbf{B}}\textbf{Limitations}\dotfill\pageref{app:limitations}

\vspace{0.8em}
% Section B
\noindent\makebox[1.5em][l]{\textbf{C}}\textbf{Model-level Dimension-wise Analysis}\dotfill\pageref{app:model_level_dimension}

\vspace{0.8em}
% Section C
\noindent\makebox[1.5em][l]{\textbf{D}}\textbf{Analysis of Scaling Effects}\dotfill\pageref{app:scaling_effects}

\vspace{0.8em}
% Section D
\noindent\makebox[1.5em][l]{\textbf{E}}\textbf{Selection of Evaluation Metrics via Human Preference Alignment}\dotfill\pageref{app:evaluation_metric_selection}

\vspace{0.8em}
% Section E
\noindent\makebox[1.5em][l]{\textbf{F}}\textbf{Benchmark Stability and Impact of Video Generators}\dotfill\pageref{app:generator_analysis}

\vspace{0.8em}
% Section F
\noindent\makebox[1.5em][l]{\textbf{G}}\textbf{Factor Coupling Analysis and Core Challenges}\dotfill\pageref{app:factor_coupling}

\vspace{0.8em}
% Section G
\noindent\makebox[1.5em][l]{\textbf{H}}\textbf{Implementation Details}\dotfill\pageref{app:implementation_details}

}

\section{Appendix Analysis}
\label{app:Appendix_Analysis}
We provide additional appendix analyses to support the main findings and improve reproducibility. Appendix~\ref{app:limitations} discusses the limitations of VI-Bench. Appendix~\ref{app:model_level_dimension} reports model-level per-dimension results to show how prompt-to-replay gaps vary across models and factors. Appendix~\ref{app:scaling_effects} studies scaling effects within the Qwen2.5-VL and Qwen3-VL families. Appendix~\ref{app:evaluation_metric_selection} explains how we select the final prompt- and video-level evaluators based on human preference alignment. Appendix~\ref{app:generator_analysis} analyzes the impact of different generators and verifies the stability of VI-Bench across generators. Appendix~\ref{app:factor_coupling} examines correlations among the five generative dimensions to reveal structured relationships between inversion abilities. Appendix~\ref{app:implementation_details} provides implementation details, API costs, benchmark examples, and system prompts.

\section{Limitations}
\label{app:limitations}

Although VI-Bench is designed to evaluate video prompt inversion in a controlled and replay-based manner, it still has several limitations. 
First, due to the high cost of repeatedly calling video generators and the difficulty of ensuring consistent replay under fixed generation settings, we do not include some of the strongest closed-source video generation models in our benchmark construction. 
Instead, we use Wan and Hunyuan as the video generators for building VI-Bench. 
This choice may introduce certain generator-specific biases, since different generators can vary in visual style, motion quality, camera behavior, and prompt-following characteristics. 
However, our generator-specific analysis in Appendix~\ref{app:generator_analysis} shows that although the choice of generator affects the absolute difficulty of inversion, especially the replay fidelity, it does not change the overall model ranking or the main findings of VI-Bench. 
This suggests that our conclusions are not artifacts of a single video generation pipeline.

Second, VI-Bench focuses on the inversion of visual content in AIGC videos. 
In this version, we do not evaluate the recovery of non-visual modalities such as subtitles, speech, sound effects, background music, or other audio information. 
However, real-world AIGC videos often contain multimodal signals, and these signals may also carry important prompt-level information. 
Future work may extend video prompt inversion beyond visual prompts by incorporating audio, speech, and text overlays, enabling a more complete evaluation of multimodal prompt recoverability.

\section{Model-level Dimension-wise Analysis}
\label{app:model_level_dimension}

To complement the aggregate dimension-wise analysis in Section~\ref{sec:dimension_analysis}, we further report model-level per-dimension results in Table~\ref{tab:overall_dimension_results}. The table provides the Prompt Score, Video Score, and Inversion Score of each evaluated model across the five generation-critical dimensions: \textit{Subject}, \textit{Action}, \textit{Scene}, \textit{Style}, and \textit{Camera}. 

The results show that frontier models exhibit the largest Prompt--Video drops on \textit{Subject} and \textit{Style}. For example, Doubao-Seed-2.0-pro drops from $3.57$ to $2.80$ on Subject and from $3.99$ to $3.09$ on Style, while GPT-4o drops from $3.41$ to $2.46$ and from $3.62$ to $2.78$, respectively. This indicates that even strong models can often identify the main subject or visual style at the text level, but fail to express them in a sufficiently precise and replay-stable form for generation. By contrast, \textit{Action} and \textit{Scene} are generally more balanced across models, suggesting that once these factors are recovered in the prompt, they are relatively easier to preserve during replay. \textit{Camera} follows a different pattern: for several weaker open-source models, the video-side score even exceeds the prompt-side score, indicating that camera information is often under-specified in the inferred prompt and only partially compensated by generator priors during replay. These results further support our main finding that video prompt inversion is not limited by uniformly weak understanding across all factors, but by factor-dependent failures in converting recovered visual information into generative control.

\begin{table*}[h]
\centering
\caption{\textbf{Overall per-dimension results on VI-Bench.} 
We report per-dimension scores for 18 VLMs. 
The \textit{Overall} columns are computed by sample-level aggregation following the main VI-Bench protocol, rather than by directly averaging the five displayed dimensions. 
\textbf{Bold} indicates the best result in each column; \underline{underline} indicates the second-best.}
\label{tab:overall_dimension_results}
\resizebox{\textwidth}{!}{%
\renewcommand{\arraystretch}{1.08}
\setlength{\tabcolsep}{4.5pt}
\begin{tabular}{l *{18}{c}}
\toprule
\multirow{2}{*}{\textbf{Method}} 
& \multicolumn{3}{c}{\cellcolor{tableheader}\textbf{Subject}} 
& \multicolumn{3}{c}{\cellcolor{tableheader}\textbf{Action}} 
& \multicolumn{3}{c}{\cellcolor{tableheader}\textbf{Scene}} 
& \multicolumn{3}{c}{\cellcolor{tableheader}\textbf{Style}} 
& \multicolumn{3}{c}{\cellcolor{tableheader}\textbf{Camera}} 
& \multicolumn{3}{c}{\cellcolor{tableheader}\textbf{Overall}} \\
\cmidrule(lr){2-4} \cmidrule(lr){5-7} \cmidrule(lr){8-10} \cmidrule(lr){11-13} \cmidrule(lr){14-16} \cmidrule(lr){17-19}
& \textbf{Prompt} & \textbf{Video} & \cellcolor{invcol}\textbf{Inv.}
& \textbf{Prompt} & \textbf{Video} & \cellcolor{invcol}\textbf{Inv.}
& \textbf{Prompt} & \textbf{Video} & \cellcolor{invcol}\textbf{Inv.}
& \textbf{Prompt} & \textbf{Video} & \cellcolor{invcol}\textbf{Inv.}
& \textbf{Prompt} & \textbf{Video} & \cellcolor{invcol}\textbf{Inv.}
& \textbf{Prompt} & \textbf{Video} & \cellcolor{invcol}\textbf{Inv.} \\
\midrule

% ============= Proprietary MLLMs =============
\rowcolor{groupshade} \multicolumn{19}{l}{\textit{\textbf{Proprietary MLLMs}}} \\

Doubao-Seed-2.0-pro
& \textbf{3.57} & \textbf{2.80} & \IC\textbf{3.18}
& \textbf{3.16} & \textbf{2.84} & \IC\textbf{3.00}
& \textbf{3.49} & \textbf{3.22} & \IC\textbf{3.36}
& \textbf{3.99} & \textbf{3.09} & \IC\textbf{3.54}
& \textbf{3.51} & \textbf{3.21} & \IC\textbf{3.36}
& \textbf{3.370} & \textbf{2.955} & \IC\textbf{3.160} \\

GPT-4o
& \underline{3.41} & 2.46 & \IC 2.94
& \underline{2.83} & \underline{2.54} & \IC\underline{2.68}
& 3.33 & \underline{2.92} & \IC 3.13
& 3.62 & \underline{2.78} & \IC\underline{3.20}
& \underline{3.13} & \underline{2.90} & \IC\underline{3.01}
& \underline{3.175} & \underline{2.635} & \IC\underline{2.905} \\

\midrule

% ============= Open-source MLLMs =============
\rowcolor{groupshade} \multicolumn{19}{l}{\textit{\textbf{Open-source MLLMs}}} \\

OmniVinci
& 3.40 & 2.45 & \IC\underline{2.95}
& 2.82 & 2.53 & \IC 2.66
& \underline{3.43} & 2.85 & \IC 3.10
& \underline{3.68} & 2.71 & \IC 3.15
& 3.06 & 2.88 & \IC 2.97
& 3.165 & 2.595 & \IC 2.880 \\

Qwen2.5-VL-72B
& 3.37 & \underline{2.49} & \IC 2.93
& 2.81 & \underline{2.55} & \IC 2.68
& 3.32 & 2.83 & \IC 3.07
& 3.59 & 2.69 & \IC 3.14
& 2.94 & 2.89 & \IC 2.92
& 3.070 & 2.620 & \IC 2.845 \\

Qwen3-VL-8B
& 3.12 & 2.48 & \IC 2.80
& 2.55 & 2.46 & \IC 2.50
& 3.08 & 2.79 & \IC 2.94
& 3.43 & 2.69 & \IC 3.06
& 2.84 & 2.77 & \IC 2.81
& 2.960 & 2.610 & \IC 2.785 \\

Qwen3-VL-30B
& 3.11 & 2.43 & \IC 2.77
& 2.51 & 2.42 & \IC 2.47
& 3.10 & 2.79 & \IC 2.94
& 3.37 & 2.66 & \IC 3.01
& 2.88 & 2.77 & \IC 2.82
& 2.940 & 2.560 & \IC 2.750 \\

LLaVA-Video
& 3.31 & 2.33 & \IC 2.82
& 2.69 & 2.38 & \IC 2.53
& 3.23 & 2.75 & \IC 2.99
& 3.45 & 2.60 & \IC 3.02
& 2.83 & 2.74 & \IC 2.78
& 3.010 & 2.470 & \IC 2.740 \\

Qwen3-VL-4B
& 3.07 & 2.43 & \IC 2.75
& 2.47 & 2.41 & \IC 2.44
& 3.01 & 2.75 & \IC 2.88
& 3.34 & 2.65 & \IC 2.99
& 2.84 & 2.78 & \IC 2.81
& 2.900 & 2.540 & \IC 2.720 \\

Qwen2.5-VL-32B
& 3.03 & 2.35 & \IC 2.69
& 2.55 & 2.42 & \IC 2.48
& 3.00 & 2.74 & \IC 2.87
& 3.27 & 2.61 & \IC 2.94
& 2.95 & 2.74 & \IC 2.84
& 2.915 & 2.495 & \IC 2.705 \\

Qwen2.5-VL-7B
& 3.27 & 2.25 & \IC 2.76
& 2.61 & 2.32 & \IC 2.47
& 3.18 & 2.58 & \IC 2.88
& 3.38 & 2.48 & \IC 2.93
& 2.55 & 2.65 & \IC 2.60
& 2.905 & 2.340 & \IC 2.620 \\

Qwen3.5
& 2.92 & 2.22 & \IC 2.57
& 2.47 & 2.22 & \IC 2.34
& 2.86 & 2.52 & \IC 2.69
& 3.23 & 2.40 & \IC 2.82
& 2.78 & 2.51 & \IC 2.64
& 2.870 & 2.320 & \IC 2.595 \\

VideoLLaMA3
& 3.20 & 2.15 & \IC 2.67
& 2.54 & 2.24 & \IC 2.39
& 3.00 & 2.48 & \IC 2.74
& 3.12 & 2.40 & \IC 2.76
& 2.51 & 2.57 & \IC 2.54
& 2.750 & 2.230 & \IC 2.490 \\

Keye-VL
& 2.65 & 1.91 & \IC 2.28
& 2.16 & 1.98 & \IC 2.07
& 2.72 & 2.39 & \IC 2.56
& 2.95 & 2.26 & \IC 2.61
& 2.51 & 2.36 & \IC 2.43
& 2.685 & 2.145 & \IC 2.415 \\

Qwen2.5-VL-3B
& 2.88 & 1.94 & \IC 2.41
& 2.23 & 2.00 & \IC 2.12
& 2.73 & 2.26 & \IC 2.50
& 2.93 & 2.17 & \IC 2.55
& 2.49 & 2.40 & \IC 2.44
& 2.550 & 2.040 & \IC 2.295 \\

InternVL3
& 2.20 & 2.16 & \IC 2.18
& 1.76 & 2.25 & \IC 2.00
& 2.11 & 2.60 & \IC 2.35
& 2.07 & 2.46 & \IC 2.27
& 1.71 & 2.60 & \IC 2.15
& 1.555 & 2.355 & \IC 1.955 \\

InternVL2.5
& 2.06 & 2.12 & \IC 2.09
& 1.61 & 2.16 & \IC 1.89
& 1.95 & 2.50 & \IC 2.22
& 1.94 & 2.38 & \IC 2.16
& 1.68 & 2.51 & \IC 2.09
& 1.470 & 2.285 & \IC 1.875 \\

PyVision-Video
& 2.64 & 0.99 & \IC 1.81
& 2.02 & 1.10 & \IC 1.56
& 2.54 & 1.40 & \IC 1.97
& 2.37 & 1.30 & \IC 1.83
& 2.06 & 1.48 & \IC 1.77
& 2.360 & 1.210 & \IC 1.785 \\

LongVideoAgent
& 2.08 & 1.58 & \IC 1.83
& 1.60 & 1.67 & \IC 1.63
& 1.87 & 1.87 & \IC 1.87
& 1.94 & 1.80 & \IC 1.87
& 1.67 & 2.07 & \IC 1.87
& 1.820 & 1.615 & \IC 1.715 \\

\bottomrule
\end{tabular}%
}
\vspace{-2pt}
\end{table*}

\section{Analysis of Scaling Effects}
\label{app:scaling_effects}

\begin{figure*}[h]
  \centering
\includegraphics[width=1.0\linewidth]{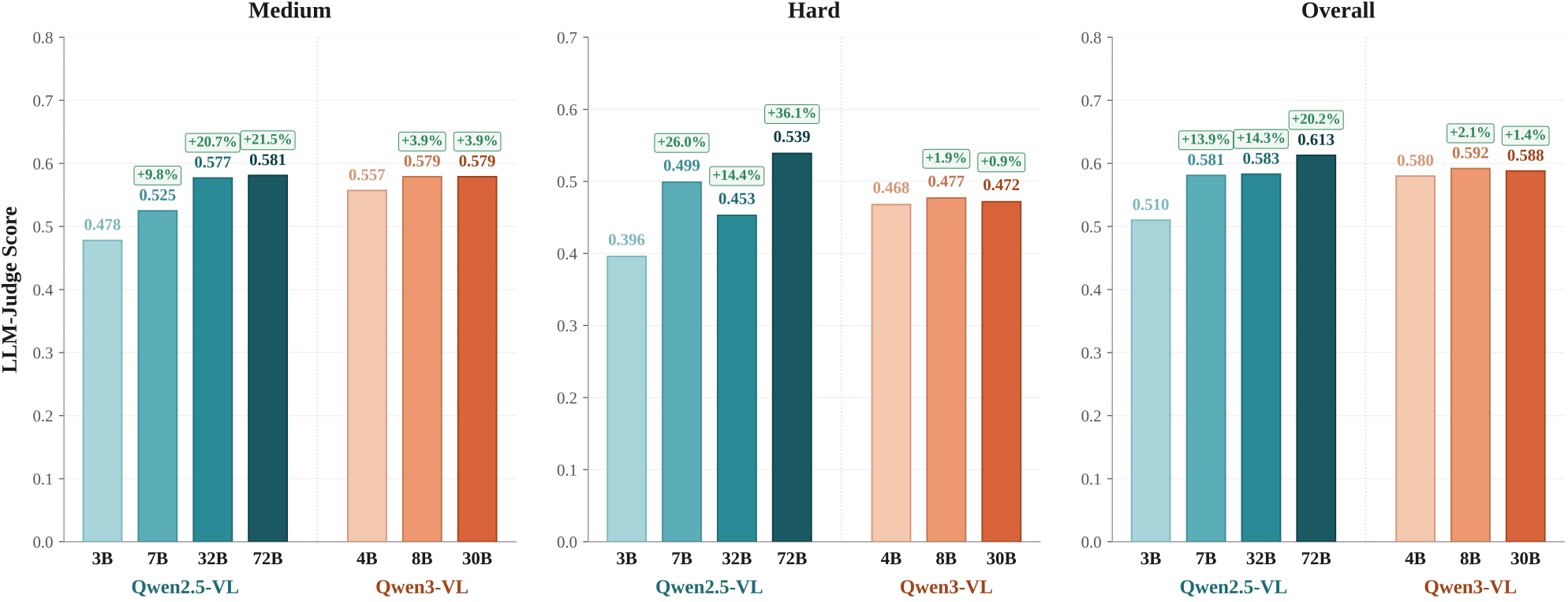}
\caption{Intra-family scaling effects on VI-Bench. Performance of Qwen2.5-VL and Qwen3-VL across model sizes under the Medium, Hard, and Overall settings.}
\label{fig:intra_family_scaling}
\end{figure*}

\begin{figure*}[h]
  \centering
\includegraphics[width=1.0\linewidth]{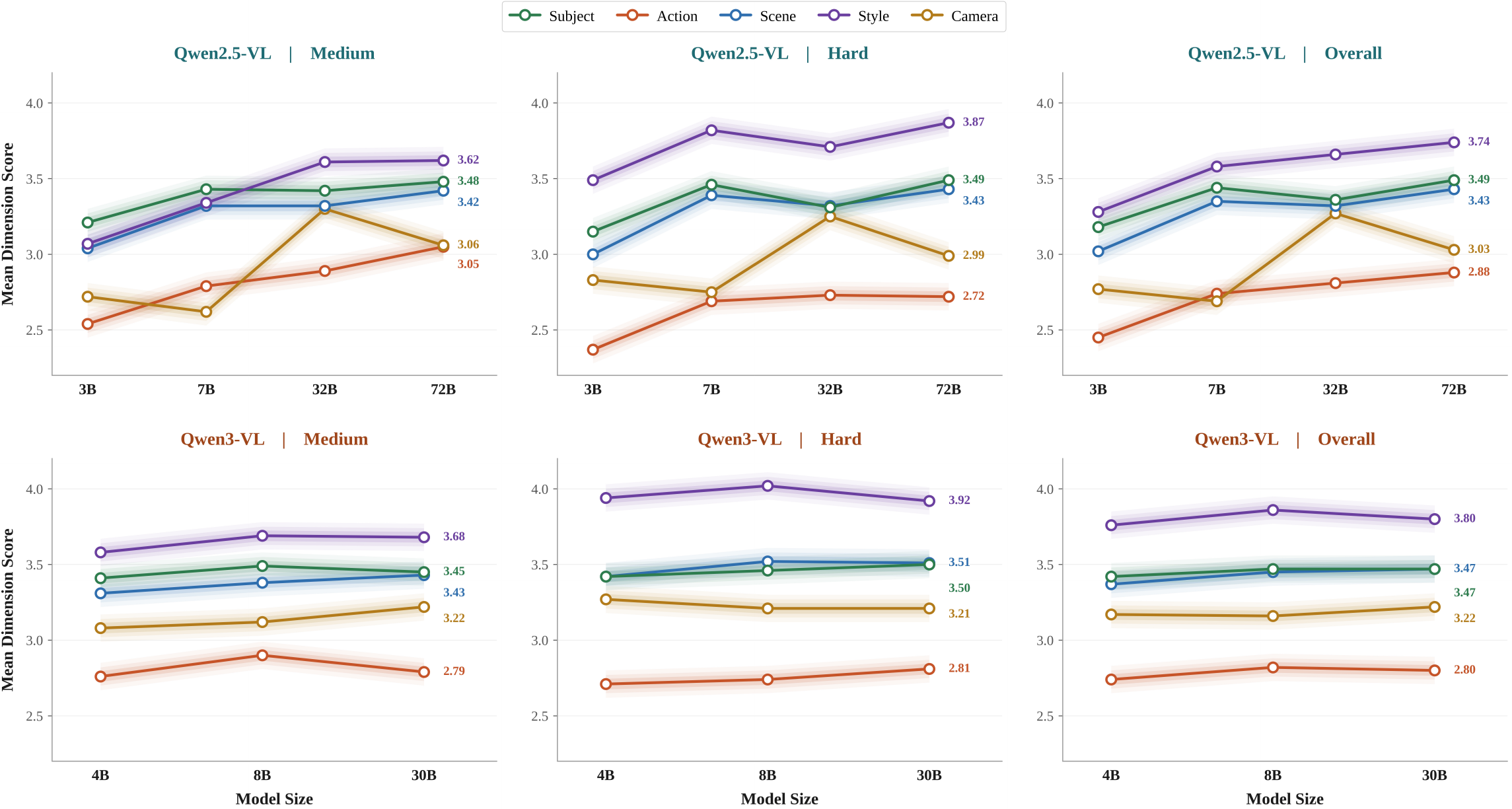}
\caption{Dimension-wise scaling trends. Per-dimension performance changes of Qwen2.5-VL and Qwen3-VL across model sizes under the Medium, Hard, and Overall settings.}
\label{fig:dimension_scores_vs_size}
\end{figure*}

To examine whether model scaling improves performance on the video prompt inversion task, we conduct a scaling analysis on two representative VLM families: Qwen2.5-VL series and Qwen3-VL series.
First, we analyze the overall scaling trend across different difficulty settings, comparing Qwen2.5-VL at four scales (3B, 7B, 32B, and 72B) and Qwen3-VL at three scales (4B, 8B, and 30B) on Medium, Hard, and Overall samples, as shown in Figure~\ref{fig:intra_family_scaling}. Second, we examine how scaling affects different generation-critical dimensions, including Subject, Action, Scene, Style, and Camera, under the same model families and difficulty settings, as shown in Figure~\ref{fig:dimension_scores_vs_size}.

The results show that scaling improves video prompt inversion, but the gains are neither uniform across model families nor evenly distributed across generative factors.  As shown in Figure~\ref{fig:intra_family_scaling}, Qwen2.5-VL exhibits clear scaling gains: its score increases from $0.478$ to $0.581$ on Medium, from $0.396$ to $0.539$ on Hard, and from $0.510$ to $0.613$ Overall, with the largest relative gain appearing on Hard samples.  This suggests that larger models are better able to handle richer control factors and multi-shot reasoning.  In contrast, Qwen3-VL starts from a stronger small model but shows much smaller marginal gains, improving only slightly from 4B to 8B/30B across all settings.  As shown in Figure~\ref{fig:dimension_scores_vs_size}, the dimension-wise trends further reveal that scaling mainly strengthens semantic and appearance-related factors such as Subject, Scene, and especially Style, while Action remains consistently lower and Camera improves only weakly.  For example, in the Overall setting, Qwen2.5-VL improves substantially in Style and Subject as the model grows, whereas Action remains below the other dimensions and Camera shows non-monotonic behavior.  Qwen3-VL shows an even more stable pattern: Style remains the strongest dimension, Subject and Scene stay relatively high, while Action is consistently the weakest.  These findings indicate that larger models become better at recognizing what should be recovered, but do not automatically acquire temporally structured action understanding or precise camera control.  

\section{Selection of Evaluation Metrics via Human Preference Alignment}
\label{app:evaluation_metric_selection}

To select reliable evaluation metrics for VI-Bench, we compare multiple candidate automatic metrics and choose the final protocol according to their agreement with human preferences.  Rather than assuming a metric a priori, we construct a human-annotated validation subset and measure how well each candidate metric correlates with human scores.  Specifically, we sample 100 benchmark instances covering different difficulty levels, VLMs, and video generators.  Each instance is independently scored by five annotators. For the video-level task, annotators are shown the ground-truth video and the replayed inversion video, and score their similarity from 1 to 5 along five aspects: subject content, scene environment, cinematography, visual style, and narrative. For the prompt-level task, annotators are shown the ground-truth prompt and the inferred prompt, and assign an overall alignment score from 1 to 5.  We average the annotator scores for each instance as the human preference reference, and then compute the Pearson Linear Correlation Coefficient (PLCC) between each automatic metric and the human scores. We evaluate six video-level metrics and two prompt-level metrics.  For video-level evaluation, the candidates include:  \textbf{Video-EvalAgent}, where a Memory Agent summarizes each video independently and a Judge Agent compares the resulting memories; \textbf{Video-Gemini}, where the ground-truth and inversion videos are jointly provided to the Gemini API for direct multi-video comparison;  \textbf{Video-PyVision}, which follows the same agent-based framework as Video-EvalAgent but replaces the Qwen3-VL Memory Agent with the agent-style VLM PyVision-Video; \textbf{Video-VLM-Concat}, where the ground-truth and inversion videos are concatenated and then evaluated by Qwen3-VL; \textbf{Video-EvalAgent-MultiShot}, which performs shot-level evaluation and averages scores across shots;  and \textbf{CLIP-I}, which computes frame-level CLIP image similarity between the two videos.  For prompt-level evaluation, we compare \textbf{LLM-Judge}, the prompt evaluator used in VI-Bench, with \textbf{CLIP-T}, which computes CLIP text embedding similarity between the ground-truth and inferred prompts. As shown in Table~\ref{tab:human_consistency}, Video-EvalAgent achieves the highest correlation with human preferences among all video-level metrics, with a PLCC of $+0.6864$. It slightly outperforms Video-Gemini ($+0.6806$) and performs better than Video-PyVision ($+0.6463$), Video-VLM-Concat ($+0.5688$), Video-EvalAgent-MultiShot ($+0.4849$), and CLIP-I ($+0.3416$). This result suggests that decomposing each video into a structured memory before comparison provides a stronger alignment with human judgments than directly concatenating videos, averaging shot-level comparisons, or using frame-level CLIP similarity. Although Video-Gemini is also competitive, Video-EvalAgent obtains the best human alignment and does not rely on direct multi-video comparison APIs. We therefore adopt Video-EvalAgent as the video-level evaluation metric in VI-Bench. For prompt-level evaluation, LLM-Judge achieves a higher correlation with human preferences than CLIP-T. Based on this human preference alignment study, we adopt LLM-Judge for prompt-level evaluation and Video-EvalAgent for video-level evaluation in the main VI-Bench protocol.

\begin{table}[t]
\centering
\caption{\textbf{Validation of evaluation metrics against human preferences.}
We compute the Pearson Linear Correlation Coefficient (PLCC) between each automatic metric and human-annotated scores on randomly sampled benchmark instances.}
\label{tab:human_consistency}
\renewcommand{\arraystretch}{1.18}
\setlength{\tabcolsep}{14pt}
\begin{tabular}{l c c}
\toprule
\rowcolor{tableheader}
\textbf{Evaluation Metric} & \textbf{PLCC ($r$)} & \textbf{Adopted} \\
\midrule
 
% ============= Video-Level Task =============
\rowcolor{taskvideo} \multicolumn{3}{l}{\textit{\textbf{Video-Level Evaluation}}} \\
 
\textbf{Video-EvalAgent} (Ours) & \cellcolor{bestcell!18}\textbf{+0.6864} & \textcolor{checkcolor}{\ding{51}} \\
Video-Gemini      & \cellcolor{secondcell!22}\underline{+0.6806} & -- \\
Video-PyVision    & +0.6463 & -- \\
Video-VLM-Concat  & +0.5688 & -- \\
Video-EvalAgent-MultiShot  & +0.4849 & -- \\
CLIP-I            & +0.3416 & -- \\
 
\midrule
% ============= Prompt-Level Task =============
\rowcolor{taskprompt} \multicolumn{3}{l}{\textit{\textbf{Prompt-Level Evaluation}}} \\
 
\textbf{LLM-Judge} (Ours) & \cellcolor{bestcell!18}\textbf{+0.6044} & \textcolor{checkcolor}{\ding{51}} \\
CLIP-T            & \cellcolor{secondcell!22}\underline{+0.4629} & -- \\
 
\bottomrule
\end{tabular}
\end{table}

% \begin{figure*}[h]
%   \centering
% \includegraphics[width=1.0\linewidth]{Figures/Video_anno1.pdf}
% \caption{Intra-family scaling effects on VI-Bench. Performance of Qwen2.5-VL and Qwen3-VL across model sizes under the Medium, Hard, and Overall settings.}
% \label{fig:intra_family_scaling}
% \end{figure*}

% \begin{figure*}[h]
%   \centering
% \includegraphics[width=1.0\linewidth]{Figures/Video_anno2.pdf}
% \caption{Intra-family scaling effects on VI-Bench. Performance of Qwen2.5-VL and Qwen3-VL across model sizes under the Medium, Hard, and Overall settings.}
% \label{fig:intra_family_scaling}
% \end{figure*}

% \begin{figure*}[h]
%   \centering
% \includegraphics[width=1.0\linewidth]{Figures/Prompt_anno.pdf}
% \caption{Intra-family scaling effects on VI-Bench. Performance of Qwen2.5-VL and Qwen3-VL across model sizes under the Medium, Hard, and Overall settings.}
% \label{fig:intra_family_scaling}
% \end{figure*}

\section{Benchmark Stability and Impact of Video Generators}
\label{app:generator_analysis}

Since VI-Bench contains samples generated by two different video generators, Wan and Hunyuan, we further analyze whether the benchmark results are sensitive to the choice of generator and examine how different generators affect video prompt inversion difficulty.

Specifically, we split VI-Bench by generator and report results on Wan-generated and Hunyuan-generated samples in Table~\ref{tab:wan_results} and Table~\ref{tab:hunyuan_results}, respectively. The results show that the choice of generator does affect the absolute difficulty of video prompt inversion, but this effect is mainly reflected in replay fidelity rather than prompt fidelity. Overall, models achieve higher Inversion Scores on Wan-generated samples than on Hunyuan-generated samples. However, this gap comes primarily from the Video Score: averaged over all models, the Prompt Score decreases only moderately from Wan to Hunyuan, whereas the Video Score drops much more substantially.  This suggests that different generators do not merely change whether a VLM can infer a semantically plausible prompt from the input video; rather, they more strongly affect whether the recovered prompt remains stable when replayed by the original generator.  In other words, generator-specific differences mainly influence the second stage of video prompt inversion, where the predicted prompt is executed again to reproduce the reference video. Despite this difference in absolute difficulty, the relative model rankings remain highly consistent across the two generators.  Strong models such as Doubao-Seed-2.0-pro, GPT-4o, OmniVinci, Qwen2.5-VL-72B, and Qwen3-VL remain among the top-performing methods on both Wan and Hunyuan, while weaker models remain near the bottom across both subsets.  This stability indicates that different generators change the replay difficulty of recovered prompts, but they do not overturn the overall capability ordering or the central conclusion that current VLMs still struggle to transform visual understanding into replay-stable generative control. 

\begin{table*}[t]
\centering
\caption{\textbf{Generator-specific results on VI-Bench using Wan.} We evaluate 18 VLMs on VI-Bench with samples generated by Wan across three difficulty levels. 
\textit{Prompt Score} measures prompt fidelity; \textit{Video Score} measures replay fidelity; 
\textit{Inversion Score} is the average of the two, reflecting a model's overall ability to recover replayable prompts from AIGC videos. \textbf{Bold} indicates the best result in each column; \underline{underline} indicates the second-best.}
\label{tab:wan_results}
\resizebox{\textwidth}{!}{%
\renewcommand{\arraystretch}{1.08}
\setlength{\tabcolsep}{5pt}
\begin{tabular}{l l c *{12}{c}}
\toprule
\multirow{2}{*}{\textbf{Method}} & \multirow{2}{*}{\textbf{Release}} & \multirow{2}{*}{\textbf{\#Params}} & \multicolumn{3}{c}{\cellcolor{tableheader}\textbf{Easy}} & \multicolumn{3}{c}{\cellcolor{tableheader}\textbf{Medium}} & \multicolumn{3}{c}{\cellcolor{tableheader}\textbf{Hard}} & \multicolumn{3}{c}{\cellcolor{tableheader}\textbf{Overall}} \\
\cmidrule(lr){4-6} \cmidrule(lr){7-9} \cmidrule(lr){10-12} \cmidrule(lr){13-15}
& & & \textbf{Prompt} & \textbf{Video} & \cellcolor{invcol}\textbf{Inv.}
    & \textbf{Prompt} & \textbf{Video} & \cellcolor{invcol}\textbf{Inv.}
    & \textbf{Prompt} & \textbf{Video} & \cellcolor{invcol}\textbf{Inv.}
    & \textbf{Prompt} & \textbf{Video} & \cellcolor{invcol}\textbf{Inv.} \\
\midrule

\rowcolor{groupshade} \multicolumn{15}{l}{\textit{\textbf{Proprietary MLLMs}}} \\

Doubao-Seed-2.0-pro & 2025-10 & -- & \textbf{0.759} & \textbf{0.782} & \IC\textbf{0.771} & \textbf{0.676} & \textbf{0.645} & \IC\textbf{0.661} & \textbf{0.651} & \textbf{0.571} & \IC\textbf{0.611} & \textbf{0.695} & \textbf{0.666} & \IC\textbf{0.681} \\
GPT-4o & 2024-08 & -- & 0.748 & 0.733 & \IC 0.741 & \underline{0.638} & 0.573 & \IC\underline{0.606} & 0.574 & \underline{0.442} & \IC\underline{0.508} & \underline{0.654} & \underline{0.583} & \IC\underline{0.618} \\

\midrule

\rowcolor{groupshade} \multicolumn{15}{l}{\textit{\textbf{Open-source MLLMs}}} \\

OmniVinci & 2025-10 & 7B & \underline{0.752} & 0.735 & \IC 0.744 & 0.603 & 0.567 & \IC 0.585 & \underline{0.604} & 0.395 & \IC 0.500 & 0.653 & 0.566 & \IC 0.609 \\
Qwen2.5-VL-72B & 2025-01 & 72B & 0.736 & 0.750 & \IC 0.743 & 0.608 & 0.575 & \IC 0.592 & 0.569 & 0.408 & \IC 0.489 & 0.638 & 0.578 & \IC 0.608 \\
Qwen3-VL-8B & 2025-09 & 8B & 0.733 & \underline{0.781} & \IC\underline{0.757} & 0.596 & \underline{0.610} & \IC 0.603 & 0.493 & 0.345 & \IC 0.419 & 0.607 & 0.579 & \IC 0.593 \\
Qwen3-VL-30B & 2025-09 & 30B & 0.739 & 0.761 & \IC 0.750 & 0.603 & 0.589 & \IC 0.596 & 0.489 & 0.352 & \IC 0.420 & 0.610 & 0.567 & \IC 0.589 \\
LLaVA-Video & 2024-10 & 7B & 0.732 & 0.734 & \IC 0.733 & 0.591 & 0.541 & \IC 0.566 & 0.538 & 0.386 & \IC 0.462 & 0.620 & 0.554 & \IC 0.587 \\
Qwen3-VL-4B & 2025-09 & 4B & 0.732 & 0.744 & \IC 0.738 & 0.594 & 0.609 & \IC 0.601 & 0.484 & 0.345 & \IC 0.414 & 0.603 & 0.566 & \IC 0.585 \\
Qwen2.5-VL-32B & 2025-03 & 32B & 0.737 & 0.747 & \IC 0.742 & 0.601 & 0.586 & \IC 0.594 & 0.488 & 0.339 & \IC 0.414 & 0.609 & 0.557 & \IC 0.583 \\
Qwen2.5-VL-7B & 2025-01 & 7B & 0.736 & 0.698 & \IC 0.717 & 0.557 & 0.528 & \IC 0.543 & 0.525 & 0.353 & \IC 0.439 & 0.606 & 0.526 & \IC 0.566 \\
Qwen3.5 & 2025-11 & 9B & 0.741 & 0.721 & \IC 0.731 & 0.632 & 0.537 & \IC 0.584 & 0.442 & 0.280 & \IC 0.361 & 0.605 & 0.513 & \IC 0.559 \\
VideoLLaMA3 & 2025-01 & 7B & 0.713 & 0.680 & \IC 0.697 & 0.523 & 0.497 & \IC 0.510 & 0.502 & 0.313 & \IC 0.408 & 0.579 & 0.497 & \IC 0.538 \\
Keye-VL & 2025-10 & 8B & 0.730 & 0.730 & \IC 0.730 & 0.589 & 0.546 & \IC 0.567 & 0.345 & 0.166 & \IC 0.255 & 0.555 & 0.480 & \IC 0.518 \\
Qwen2.5-VL-3B & 2025-01 & 3B & 0.679 & 0.667 & \IC 0.673 & 0.496 & 0.444 & \IC 0.470 & 0.414 & 0.235 & \IC 0.324 & 0.530 & 0.448 & \IC 0.489 \\
InternVL3 & 2025-04 & 8B & 0.400 & 0.731 & \IC 0.565 & 0.298 & 0.528 & \IC 0.413 & 0.265 & 0.305 & \IC 0.285 & 0.321 & 0.521 & \IC 0.421 \\
InternVL2.5 & 2024-12 & 8B & 0.393 & 0.733 & \IC 0.563 & 0.300 & 0.560 & \IC 0.430 & 0.222 & 0.231 & \IC 0.226 & 0.305 & 0.508 & \IC 0.406 \\
PyVision-Video & 2025-08 & 7B & 0.672 & 0.611 & \IC 0.641 & 0.482 & 0.377 & \IC 0.429 & 0.318 & -0.198 & \IC 0.060 & 0.490 & 0.263 & \IC 0.377 \\
LongVideoAgent & 2025-05 & 7B & 0.558 & 0.580 & \IC 0.569 & 0.404 & 0.344 & \IC 0.374 & 0.173 & 0.135 & \IC 0.154 & 0.378 & 0.353 & \IC 0.366 \\

\bottomrule
\end{tabular}%
}
\vspace{-12pt}
\end{table*}

\begin{table*}[t]
\centering
\caption{\textbf{Generator-specific results on VI-Bench using Hunyuan.} We evaluate 18 VLMs on VI-Bench with samples generated by Hunyuan across three difficulty levels. 
\textit{Prompt Score} measures prompt fidelity; \textit{Video Score} measures replay fidelity; 
\textit{Inversion Score} is the average of the two, reflecting a model's overall ability to recover replayable prompts from AIGC videos. \textbf{Bold} indicates the best result in each column; \underline{underline} indicates the second-best.}
\label{tab:hunyuan_results}
\resizebox{\textwidth}{!}{%
\renewcommand{\arraystretch}{1.08}
\setlength{\tabcolsep}{5pt}
\begin{tabular}{l l c *{12}{c}}
\toprule
\multirow{2}{*}{\textbf{Method}} & \multirow{2}{*}{\textbf{Release}} & \multirow{2}{*}{\textbf{\#Params}} & \multicolumn{3}{c}{\cellcolor{tableheader}\textbf{Easy}} & \multicolumn{3}{c}{\cellcolor{tableheader}\textbf{Medium}} & \multicolumn{3}{c}{\cellcolor{tableheader}\textbf{Hard}} & \multicolumn{3}{c}{\cellcolor{tableheader}\textbf{Overall}} \\
\cmidrule(lr){4-6} \cmidrule(lr){7-9} \cmidrule(lr){10-12} \cmidrule(lr){13-15}
& & & \textbf{Prompt} & \textbf{Video} & \cellcolor{invcol}\textbf{Inv.}
    & \textbf{Prompt} & \textbf{Video} & \cellcolor{invcol}\textbf{Inv.}
    & \textbf{Prompt} & \textbf{Video} & \cellcolor{invcol}\textbf{Inv.}
    & \textbf{Prompt} & \textbf{Video} & \cellcolor{invcol}\textbf{Inv.} \\
\midrule

\rowcolor{groupshade} \multicolumn{15}{l}{\textit{\textbf{Proprietary MLLMs}}} \\

Doubao-Seed-2.0-pro & 2025-10 & -- & \underline{0.735} & \textbf{0.728} & \IC\textbf{0.732} & \textbf{0.621} & \textbf{0.523} & \IC\textbf{0.572} & \textbf{0.603} & \textbf{0.293} & \IC\textbf{0.448} & \textbf{0.653} & \textbf{0.514} & \IC\textbf{0.584} \\
GPT-4o & 2024-08 & -- & \textbf{0.736} & 0.708 & \IC\underline{0.722} & \underline{0.583} & 0.483 & \IC 0.533 & 0.527 & 0.219 & \IC 0.373 & \underline{0.615} & 0.470 & \IC\underline{0.543} \\

\midrule

\rowcolor{groupshade} \multicolumn{15}{l}{\textit{\textbf{Open-source MLLMs}}} \\

OmniVinci & 2025-10 & 7B & 0.710 & 0.691 & \IC 0.701 & 0.571 & 0.483 & \IC 0.527 & \underline{0.561} & \underline{0.240} & \IC\underline{0.401} & 0.614 & \underline{0.471} & \IC 0.543 \\
Qwen2.5-VL-72B & 2025-01 & 72B & 0.705 & 0.702 & \IC 0.703 & 0.554 & 0.490 & \IC 0.522 & 0.509 & 0.217 & \IC 0.363 & 0.589 & 0.469 & \IC 0.529 \\
Qwen3-VL-8B & 2025-09 & 8B & 0.707 & \underline{0.711} & \IC 0.709 & 0.561 & \underline{0.518} & \IC\underline{0.539} & 0.460 & 0.165 & \IC 0.312 & 0.576 & 0.465 & \IC 0.520 \\
Qwen3-VL-30B & 2025-09 & 30B & 0.684 & 0.696 & \IC 0.690 & 0.554 & 0.508 & \IC 0.531 & 0.455 & 0.164 & \IC 0.310 & 0.565 & 0.456 & \IC 0.510 \\
LLaVA-Video & 2024-10 & 7B & 0.713 & 0.670 & \IC 0.692 & 0.530 & 0.428 & \IC 0.479 & 0.509 & 0.203 & \IC 0.356 & 0.584 & 0.434 & \IC 0.509 \\
Qwen3-VL-4B & 2025-09 & 4B & 0.698 & 0.696 & \IC 0.697 & 0.520 & 0.496 & \IC 0.508 & 0.452 & 0.155 & \IC 0.304 & 0.557 & 0.449 & \IC 0.503 \\
Qwen2.5-VL-32B & 2025-03 & 32B & 0.699 & 0.675 & \IC 0.687 & 0.553 & 0.485 & \IC 0.519 & 0.417 & 0.161 & \IC 0.289 & 0.556 & 0.441 & \IC 0.498 \\
Qwen2.5-VL-7B & 2025-01 & 7B & 0.699 & 0.653 & \IC 0.676 & 0.493 & 0.411 & \IC 0.452 & 0.474 & 0.164 & \IC 0.319 & 0.555 & 0.409 & \IC 0.482 \\
Qwen3.5 & 2025-11 & 9B & 0.700 & 0.690 & \IC 0.695 & 0.559 & 0.469 & \IC 0.514 & 0.373 & 0.087 & \IC 0.230 & 0.544 & 0.415 & \IC 0.480 \\
VideoLLaMA3 & 2025-01 & 7B & 0.659 & 0.629 & \IC 0.644 & 0.437 & 0.389 & \IC 0.413 & 0.465 & 0.169 & \IC 0.317 & 0.520 & 0.396 & \IC 0.458 \\
Keye-VL & 2025-10 & 8B & 0.693 & 0.663 & \IC 0.678 & 0.547 & 0.433 & \IC 0.490 & 0.317 & 0.037 & \IC 0.177 & 0.519 & 0.378 & \IC 0.448 \\
Qwen2.5-VL-3B & 2025-01 & 3B & 0.633 & 0.628 & \IC 0.630 & 0.460 & 0.363 & \IC 0.412 & 0.378 & 0.114 & \IC 0.246 & 0.490 & 0.368 & \IC 0.429 \\
InternVL3 & 2025-04 & 8B & 0.384 & 0.683 & \IC 0.533 & 0.281 & 0.445 & \IC 0.363 & 0.240 & 0.134 & \IC 0.187 & 0.301 & 0.420 & \IC 0.361 \\
InternVL2.5 & 2024-12 & 8B & 0.373 & 0.673 & \IC 0.523 & 0.270 & 0.431 & \IC 0.350 & 0.203 & 0.111 & \IC 0.157 & 0.282 & 0.405 & \IC 0.344 \\
PyVision-Video & 2025-08 & 7B & 0.640 & 0.590 & \IC 0.615 & 0.441 & 0.286 & \IC 0.363 & 0.277 & -0.212 & \IC 0.032 & 0.453 & 0.221 & \IC 0.337 \\
LongVideoAgent & 2025-05 & 7B & 0.518 & 0.562 & \IC 0.540 & 0.365 & 0.257 & \IC 0.311 & 0.163 & 0.059 & \IC 0.111 & 0.349 & 0.293 & \IC 0.321 \\

\bottomrule
\end{tabular}%
}
\vspace{-12pt}
\end{table*}

\section{Factor Coupling Analysis and Core Challenges}
\label{app:factor_coupling}

\begin{figure*}[h]
  \centering
   % \vspace{-10pt}  % 在caption后添加，缩小与正文的间距
\includegraphics[width=1.0\linewidth]{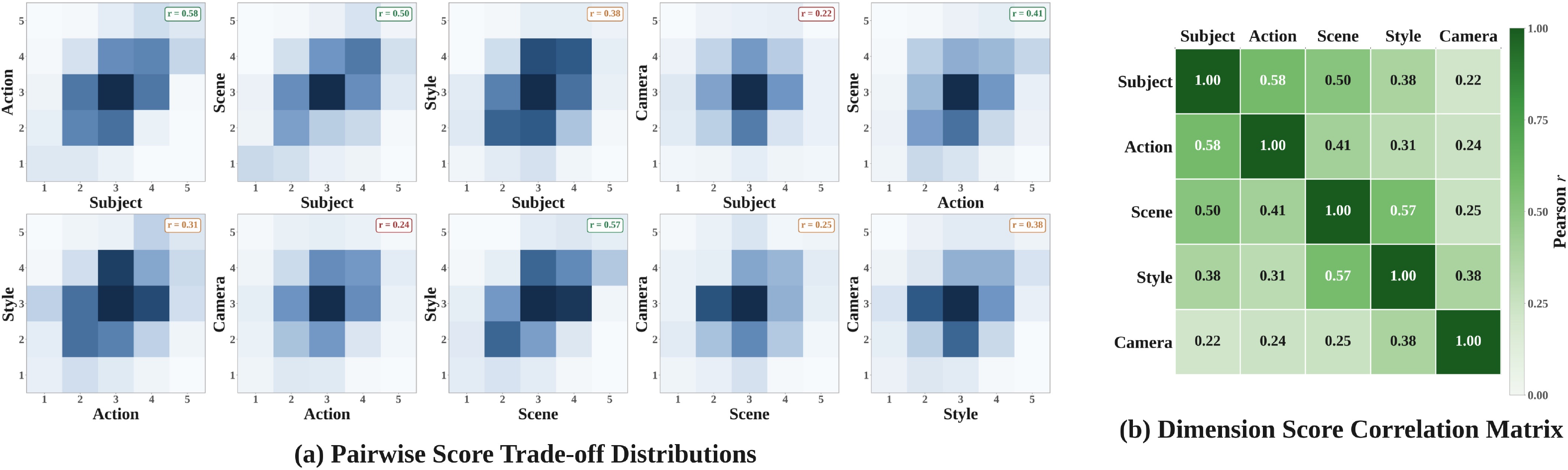}
\caption{Pairwise score trade-off distributions between dimension pairs (left) and the correlation matrix of dimension scores across the five generative factors (right).}
\label{fig:factor_analyse}
\vspace{-12pt}  % 在caption后添加，缩小与正文的间距
\end{figure*}

To investigate why video prompt inversion remains difficult even when some individual factors are partially recoverable, we further analyze the structural relationships among the five generative dimensions. As shown in Figure~\ref{fig:factor_analyse}(a) and Figure~\ref{fig:factor_analyse}(b), \textit{Subject}, \textit{Action}, and \textit{Scene} form a clear content cluster, with relatively high pairwise correlations, suggesting that recognizing what is in the video, what it is doing, and where it happens largely relies on a shared semantic understanding ability. \textit{Scene} and \textit{Style} are also strongly coupled ($r = 0.57$), indicating that style perception is often grounded in background cues such as lighting, color tone, and atmosphere. In contrast, \textit{Camera} is the only isolated dimension: its correlations with all other factors are consistently the lowest, implying that camera understanding does not naturally improve together with content understanding. This further suggests that video prompt inversion is not a task of recovering five independent labels, but of operating in a structured factor space where some abilities are mutually supportive while others are fundamentally distinct. Since \textit{Subject}, \textit{Action}, and \textit{Scene} are strongly coupled, improvements in general semantic video understanding may benefit these content-related dimensions together. However, \textit{Camera} behaves as a more independent ability: a model can become stronger at recognizing subjects, actions, and scenes while still failing to recover shot scale, viewpoint, or camera motion. \textit{Style} lies between these two cases. Its strong correlation with \textit{Scene} suggests that style is often inferred from static visual cues such as lighting, color tone, and background atmosphere, while its weaker correlation with \textit{Action} suggests that style recovery is less tied to dynamic event understanding. These observations indicate that improving video prompt inversion may require not only stronger general video understanding, but also more knowledge like camera language or styles.

\section{Implementation Details}
\label{app:implementation_details}

All experiments are conducted on a server equipped with four NVIDIA RTX 6000 GPUs. For GPT-4o, we use the \texttt{gpt-4o-2024-11-20} version throughout our evaluation. In terms of API cost, evaluating Prompt Score with LLM-Judge (GPT-4o) costs approximately \$3 for one full pass over VI-Bench. For model inference, running GPT-4o on the full VI-Bench costs approximately \$42, while running Doubao-Seed-2.0-Pro costs approximately \$8. These costs only refer to the corresponding inference or evaluation calls and may vary with API pricing or implementation details.

\begin{figure*}[h]
  \centering
\includegraphics[width=0.7\linewidth]{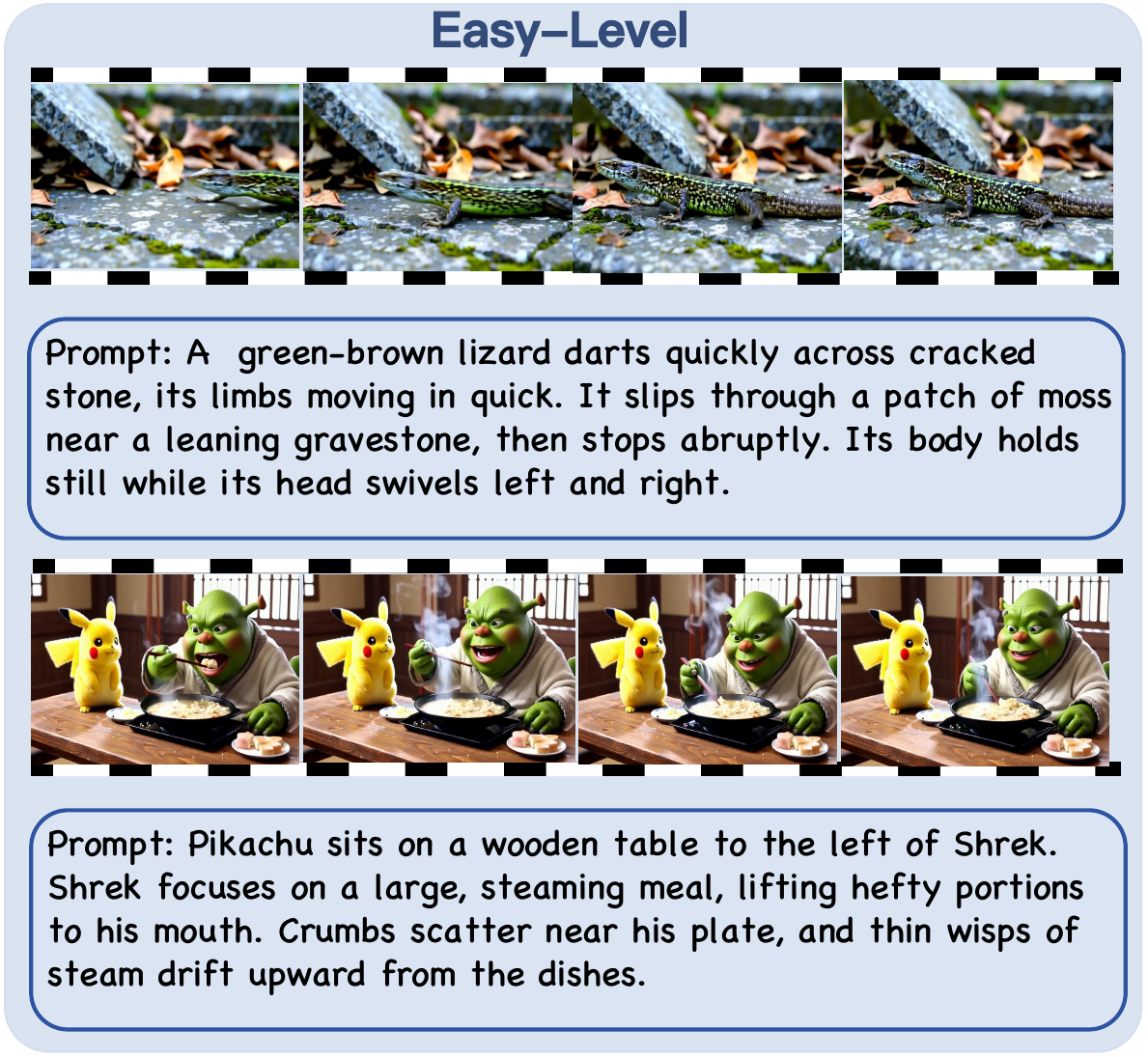}
\caption{Easy-level samples in VI-Bench.}
\label{fig:bench_figure_1}
\vspace{-12pt}
\end{figure*}

\begin{figure*}[h]
  \centering
\includegraphics[width=0.7\linewidth]{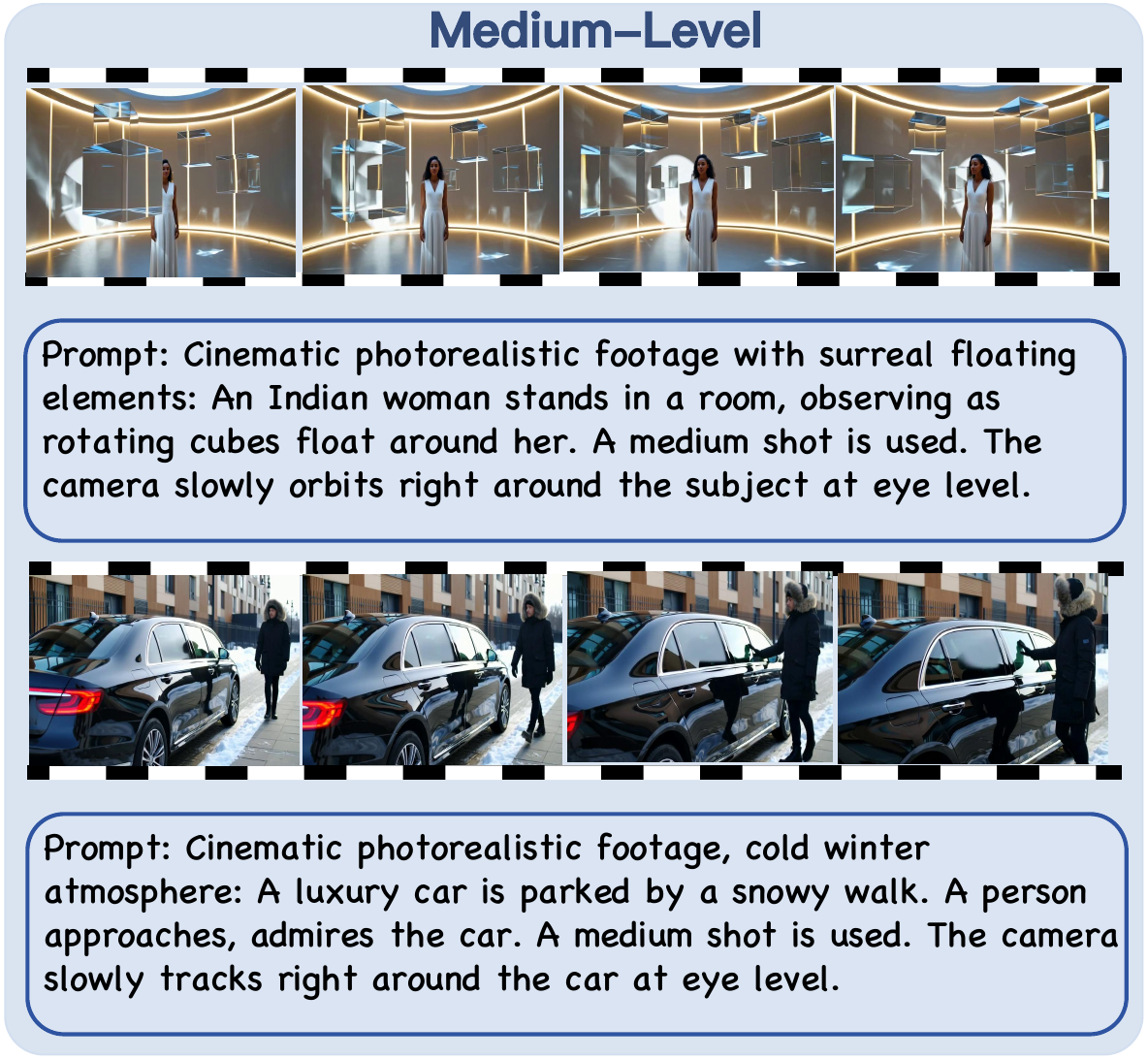}
\caption{Medium-level samples in VI-Bench.}
\label{fig:bench_figure_2}
\vspace{-12pt}
\end{figure*}

\begin{figure*}[h]
  \centering
\includegraphics[width=1.0\linewidth]{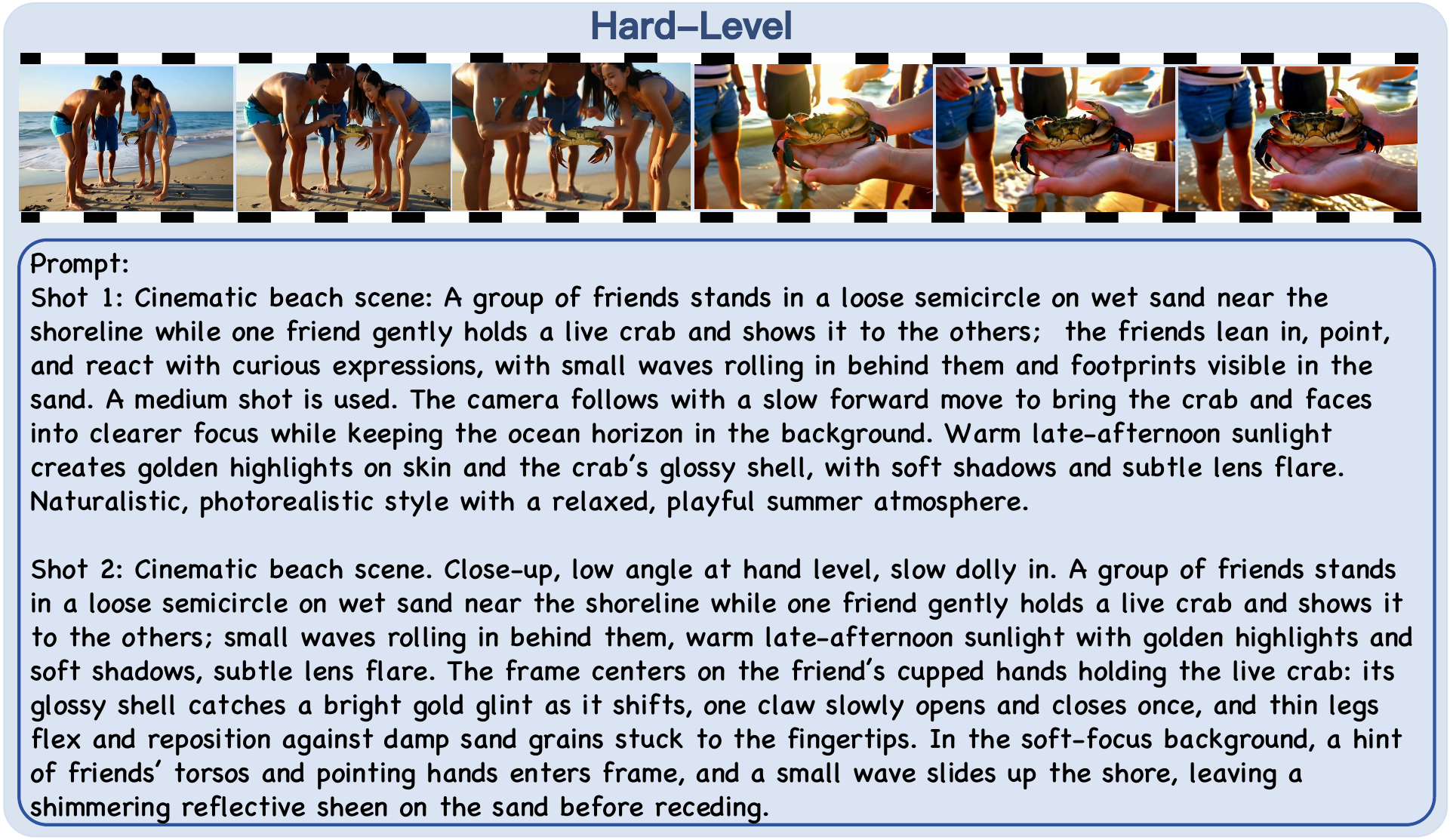}
\caption{A Hard-level multi-shot sample in VI-Bench.}
\label{fig:bench_figure_3}
\vspace{-12pt}
\end{figure*}

\begin{figure*}[h]
  \centering
\includegraphics[width=1.0\linewidth]{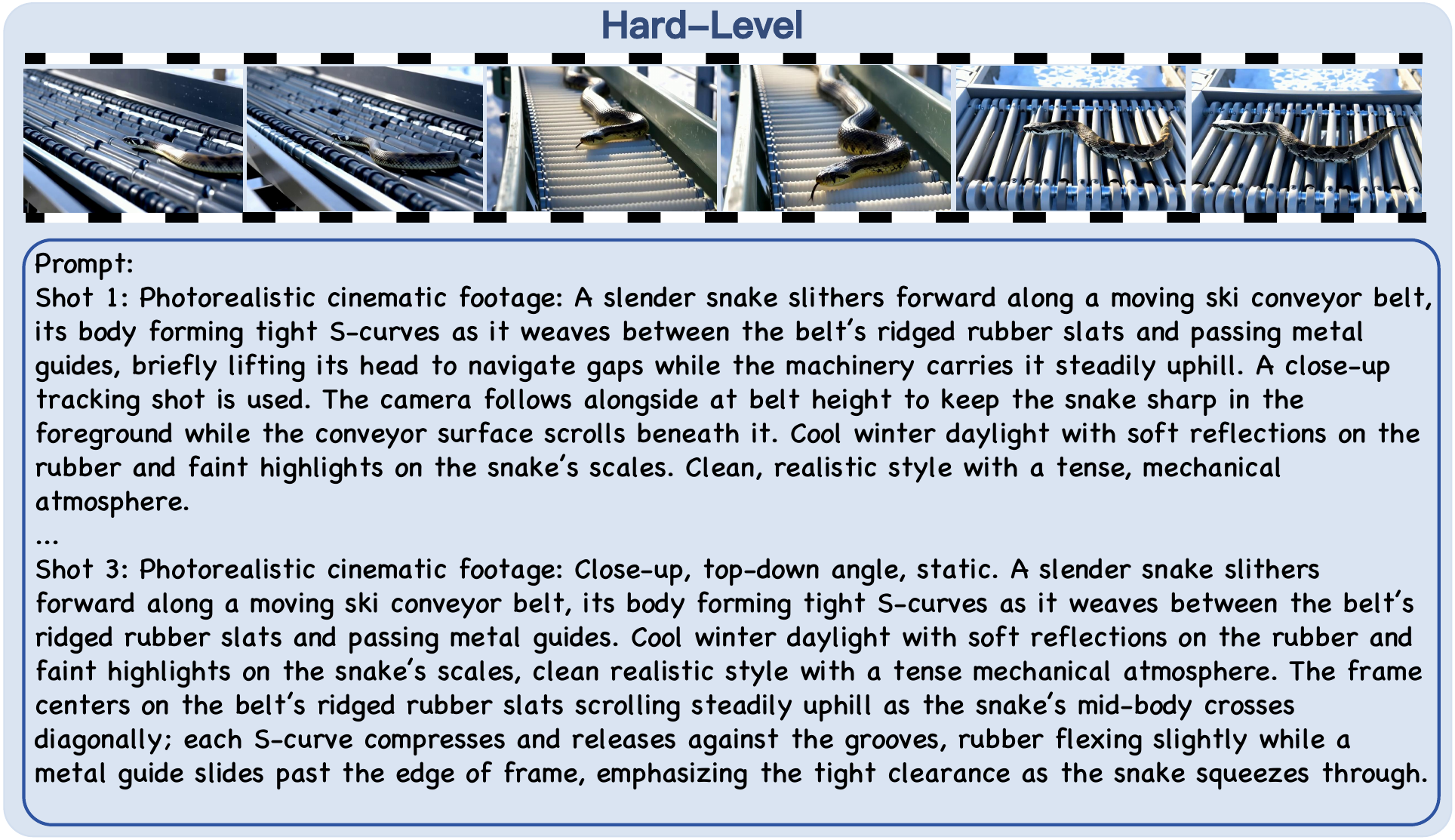}
\caption{A Hard-level multi-shot sample in VI-Bench.}
\label{fig:bench_figure_4}
\vspace{-12pt}
\end{figure*}

\begin{figure*}[h]
  \centering
\includegraphics[width=1.0\linewidth]{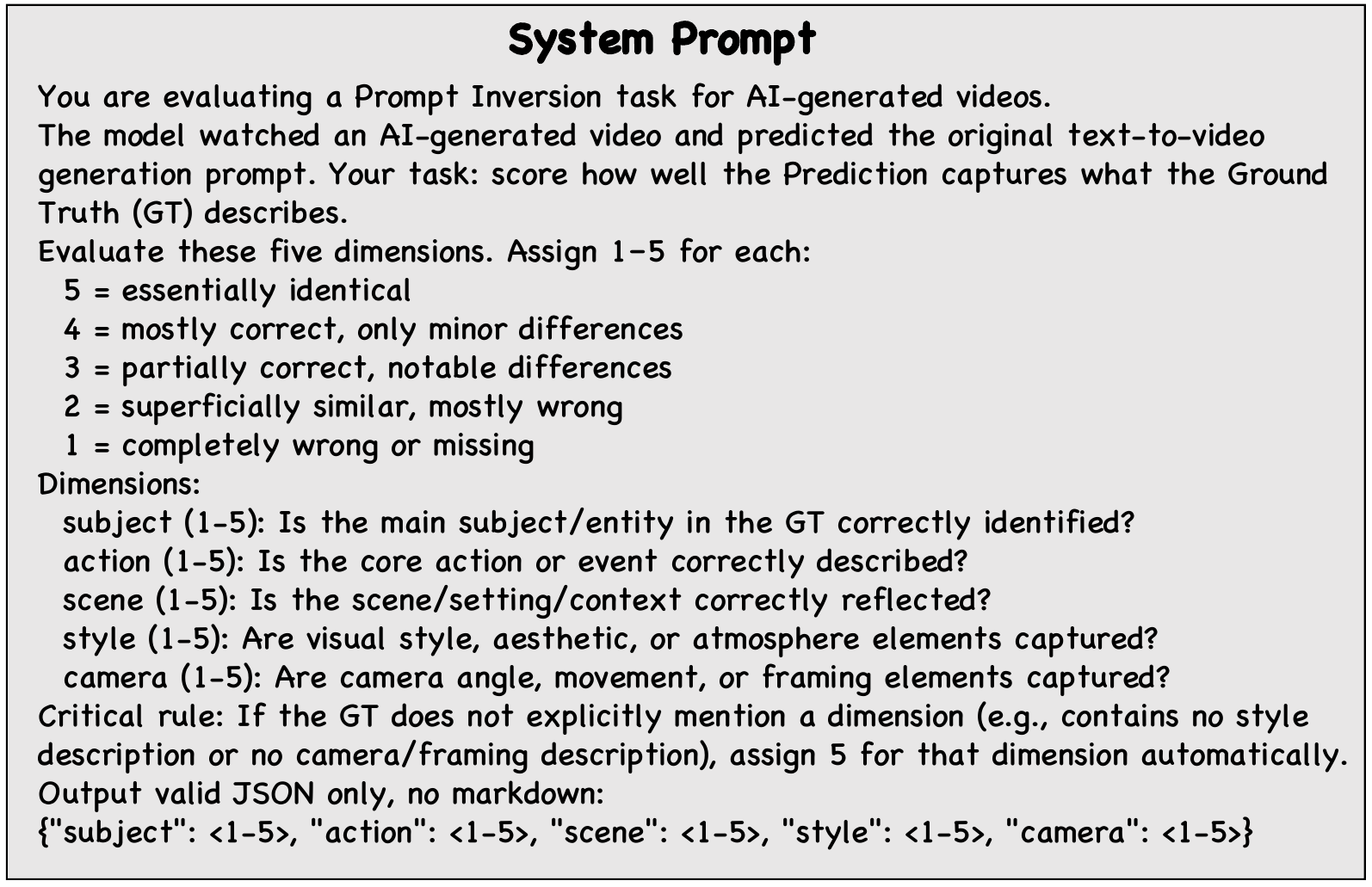}
\caption{System prompt used for Prompt Score evaluation.}
\label{fig:bench_figure_5}
\vspace{-12pt}
\end{figure*}

\begin{figure*}[h]
  \centering
\includegraphics[width=1.0\linewidth]{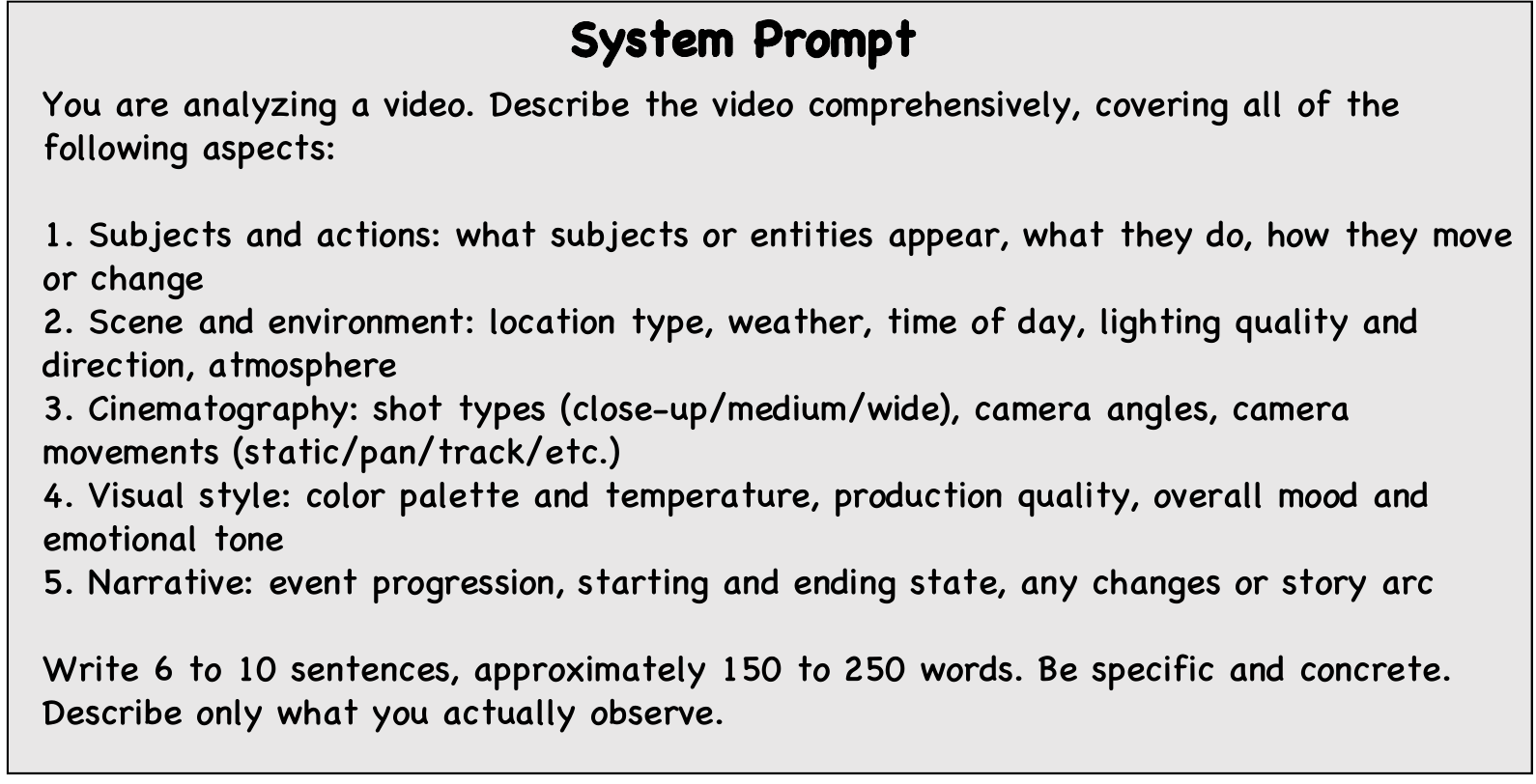}
\caption{System prompt used for the Memory Agent.}
\label{fig:bench_figure_6}
\vspace{-12pt}
\end{figure*}

\begin{figure*}[h]
  \centering
\includegraphics[width=1.0\linewidth]{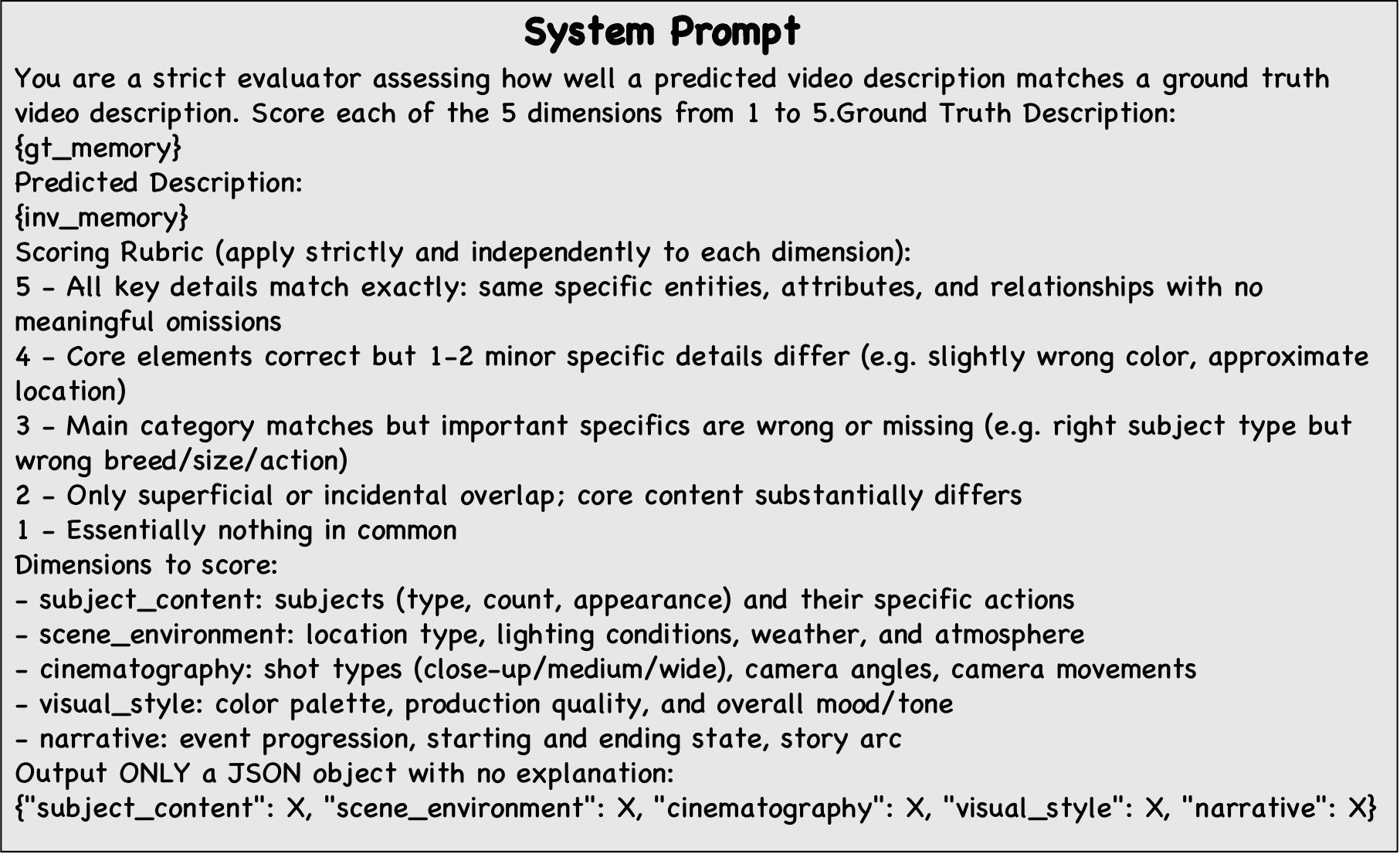}
\caption{System prompt used for the Evaluation Agent.}
\label{fig:bench_figure_7}
\vspace{-12pt}
\end{figure*}

We also provide representative examples and system prompts used in VI-Bench. Figures~\ref{fig:bench_figure_1} and~\ref{fig:bench_figure_2} show Easy- and Medium-level samples, respectively, while Figures~\ref{fig:bench_figure_3} and~\ref{fig:bench_figure_4} show Hard-level multi-shot samples. To improve reproducibility, we further present the system prompt used for Prompt Score evaluation in Figure~\ref{fig:bench_figure_5}, the system prompt used by the Memory Agent in Figure~\ref{fig:bench_figure_6}, and the system prompt used by the Evaluation Agent in Figure~\ref{fig:bench_figure_7}.

\clearpage
\newpage
\section*{NeurIPS Paper Checklist}

\begin{enumerate}

\item {\bf Claims}
    \item[] Question: Do the main claims made in the abstract and introduction accurately reflect the paper's contributions and scope?
    \item[] Answer: \answerYes{}
    \item[] Justification: The main claims in the abstract and introduction are aligned with the scope of the paper: defining video prompt inversion, introducing VI-Bench, and evaluating current VLMs under a closed-loop replay protocol. The empirical claims are supported by the benchmark construction, experimental results, and analysis sections.
    \item[] Guidelines:
    \begin{itemize}
        \item The answer \answerNA{} means that the abstract and introduction do not include the claims made in the paper.
        \item The abstract and/or introduction should clearly state the claims made, including the contributions made in the paper and important assumptions and limitations. A \answerNo{} or \answerNA{} answer to this question will not be perceived well by the reviewers. 
        \item The claims made should match theoretical and experimental results, and reflect how much the results can be expected to generalize to other settings. 
        \item It is fine to include aspirational goals as motivation as long as it is clear that these goals are not attained by the paper. 
    \end{itemize}

\item {\bf Limitations}
    \item[] Question: Does the paper discuss the limitations of the work performed by the authors?
    \item[] Answer: \answerYes{}
    \item[] Justification: The paper discusses limitations related to the benchmark scope, the limited set of video generators, the dependence on fixed generation settings, prompt non-uniqueness, and possible biases of LLM-based evaluation. These limitations are described in the limitations and broader discussion sections.
    \item[] Guidelines:
    \begin{itemize}
        \item The answer \answerNA{} means that the paper has no limitation while the answer \answerNo{} means that the paper has limitations, but those are not discussed in the paper. 
        \item The authors are encouraged to create a separate ``Limitations'' section in their paper.
        \item The paper should point out any strong assumptions and how robust the results are to violations of these assumptions (e.g., independence assumptions, noiseless settings, model well-specification, asymptotic approximations only holding locally). The authors should reflect on how these assumptions might be violated in practice and what the implications would be.
        \item The authors should reflect on the scope of the claims made, e.g., if the approach was only tested on a few datasets or with a few runs. In general, empirical results often depend on implicit assumptions, which should be articulated.
        \item The authors should reflect on the factors that influence the performance of the approach. For example, a facial recognition algorithm may perform poorly when image resolution is low or images are taken in low lighting. Or a speech-to-text system might not be used reliably to provide closed captions for online lectures because it fails to handle technical jargon.
        \item The authors should discuss the computational efficiency of the proposed algorithms and how they scale with dataset size.
        \item If applicable, the authors should discuss possible limitations of their approach to address problems of privacy and fairness.
        \item While the authors might fear that complete honesty about limitations might be used by reviewers as grounds for rejection, a worse outcome might be that reviewers discover limitations that aren't acknowledged in the paper. The authors should use their best judgment and recognize that individual actions in favor of transparency play an important role in developing norms that preserve the integrity of the community. Reviewers will be specifically instructed to not penalize honesty concerning limitations.
    \end{itemize}

\item {\bf Theory assumptions and proofs}
    \item[] Question: For each theoretical result, does the paper provide the full set of assumptions and a complete (and correct) proof?
    \item[] Answer: \answerNA{}
    \item[] Justification: The paper does not present theoretical results, theorems, lemmas, or formal proofs. Its contributions are centered on task formulation, benchmark construction, empirical evaluation, and analysis.
    \item[] Guidelines:
    \begin{itemize}
        \item The answer \answerNA{} means that the paper does not include theoretical results. 
        \item All the theorems, formulas, and proofs in the paper should be numbered and cross-referenced.
        \item All assumptions should be clearly stated or referenced in the statement of any theorems.
        \item The proofs can either appear in the main paper or the supplemental material, but if they appear in the supplemental material, the authors are encouraged to provide a short proof sketch to provide intuition. 
        \item Inversely, any informal proof provided in the core of the paper should be complemented by formal proofs provided in appendix or supplemental material.
        \item Theorems and Lemmas that the proof relies upon should be properly referenced. 
    \end{itemize}

\item {\bf Experimental result reproducibility}
    \item[] Question: Does the paper fully disclose all the information needed to reproduce the main experimental results of the paper to the extent that it affects the main claims and/or conclusions of the paper (regardless of whether the code and data are provided or not)?
    \item[] Answer: \answerYes{}
    \item[] Justification: The paper provides the benchmark construction procedure, data sources, difficulty design, model list, inference protocol, replay setting, fixed generation configuration, and scoring procedure. Additional implementation details, evaluation prompts, and reproducibility instructions are provided in the appendix and supplementary materials.
    \item[] Guidelines:
    \begin{itemize}
        \item The answer \answerNA{} means that the paper does not include experiments.
        \item If the paper includes experiments, a \answerNo{} answer to this question will not be perceived well by the reviewers: Making the paper reproducible is important, regardless of whether the code and data are provided or not.
        \item If the contribution is a dataset and\slash or model, the authors should describe the steps taken to make their results reproducible or verifiable. 
        \item Depending on the contribution, reproducibility can be accomplished in various ways. For example, if the contribution is a novel architecture, describing the architecture fully might suffice, or if the contribution is a specific model and empirical evaluation, it may be necessary to either make it possible for others to replicate the model with the same dataset, or provide access to the model. In general. releasing code and data is often one good way to accomplish this, but reproducibility can also be provided via detailed instructions for how to replicate the results, access to a hosted model (e.g., in the case of a large language model), releasing of a model checkpoint, or other means that are appropriate to the research performed.
        \item While NeurIPS does not require releasing code, the conference does require all submissions to provide some reasonable avenue for reproducibility, which may depend on the nature of the contribution. For example
        \begin{enumerate}
            \item If the contribution is primarily a new algorithm, the paper should make it clear how to reproduce that algorithm.
            \item If the contribution is primarily a new model architecture, the paper should describe the architecture clearly and fully.
            \item If the contribution is a new model (e.g., a large language model), then there should either be a way to access this model for reproducing the results or a way to reproduce the model (e.g., with an open-source dataset or instructions for how to construct the dataset).
            \item We recognize that reproducibility may be tricky in some cases, in which case authors are welcome to describe the particular way they provide for reproducibility. In the case of closed-source models, it may be that access to the model is limited in some way (e.g., to registered users), but it should be possible for other researchers to have some path to reproducing or verifying the results.
        \end{enumerate}
    \end{itemize}

\item {\bf Open access to data and code}
    \item[] Question: Does the paper provide open access to the data and code, with sufficient instructions to faithfully reproduce the main experimental results, as described in supplemental material?
    \item[] Answer: \answerYes{}
    \item[] Justification: The paper provides open access to the benchmark assets and evaluation code through anonymized supplementary materials or an anonymized repository. The release includes data documentation, evaluation scripts, prompt templates, and instructions for reproducing the main results.
    \item[] Guidelines:
    \begin{itemize}
        \item The answer \answerNA{} means that paper does not include experiments requiring code.
        \item Please see the NeurIPS code and data submission guidelines (\url{https://neurips.cc/public/guides/CodeSubmissionPolicy}) for more details.
        \item While we encourage the release of code and data, we understand that this might not be possible, so \answerNo{} is an acceptable answer. Papers cannot be rejected simply for not including code, unless this is central to the contribution (e.g., for a new open-source benchmark).
        \item The instructions should contain the exact command and environment needed to run to reproduce the results. See the NeurIPS code and data submission guidelines (\url{https://neurips.cc/public/guides/CodeSubmissionPolicy}) for more details.
        \item The authors should provide instructions on data access and preparation, including how to access the raw data, preprocessed data, intermediate data, and generated data, etc.
        \item The authors should provide scripts to reproduce all experimental results for the new proposed method and baselines. If only a subset of experiments are reproducible, they should state which ones are omitted from the script and why.
        \item At submission time, to preserve anonymity, the authors should release anonymized versions (if applicable).
        \item Providing as much information as possible in supplemental material (appended to the paper) is recommended, but including URLs to data and code is permitted.
    \end{itemize}

\item {\bf Experimental setting/details}
    \item[] Question: Does the paper specify all the training and test details (e.g., data splits, hyperparameters, how they were chosen, type of optimizer) necessary to understand the results?
    \item[] Answer: \answerYes{}
    \item[] Justification: The paper specifies the experimental setting, including data construction, difficulty levels, evaluated models, frame sampling strategy, model prompting protocol, replay configuration, scoring dimensions, and aggregation rules. Full implementation and evaluation details are provided in the appendix and supplementary materials.
    \item[] Guidelines:
    \begin{itemize}
        \item The answer \answerNA{} means that the paper does not include experiments.
        \item The experimental setting should be presented in the core of the paper to a level of detail that is necessary to appreciate the results and make sense of them.
        \item The full details can be provided either with the code, in appendix, or as supplemental material.
    \end{itemize}

\item {\bf Experiment statistical significance}
    \item[] Question: Does the paper report error bars suitably and correctly defined or other appropriate information about the statistical significance of the experiments?
    \item[] Answer: \answerNo{}
    \item[] Justification: The paper reports aggregate benchmark scores and human validation results, but does not provide error bars or confidence intervals for all main experimental results. This is mainly due to the high cost of closed-loop video replay and LLM-based evaluation across many models and samples.
    \item[] Guidelines:
    \begin{itemize}
        \item The answer \answerNA{} means that the paper does not include experiments.
        \item The authors should answer \answerYes{} if the results are accompanied by error bars, confidence intervals, or statistical significance tests, at least for the experiments that support the main claims of the paper.
        \item The factors of variability that the error bars are capturing should be clearly stated (for example, train/test split, initialization, random drawing of some parameter, or overall run with given experimental conditions).
        \item The method for calculating the error bars should be explained (closed form formula, call to a library function, bootstrap, etc.)
        \item The assumptions made should be given (e.g., Normally distributed errors).
        \item It should be clear whether the error bar is the standard deviation or the standard error of the mean.
        \item It is OK to report 1-sigma error bars, but one should state it. The authors should preferably report a 2-sigma error bar than state that they have a 96\% CI, if the hypothesis of Normality of errors is not verified.
        \item For asymmetric distributions, the authors should be careful not to show in tables or figures symmetric error bars that would yield results that are out of range (e.g., negative error rates).
        \item If error bars are reported in tables or plots, the authors should explain in the text how they were calculated and reference the corresponding figures or tables in the text.
    \end{itemize}

\item {\bf Experiments compute resources}
    \item[] Question: For each experiment, does the paper provide sufficient information on the computer resources (type of compute workers, memory, time of execution) needed to reproduce the experiments?
    \item[] Answer: \answerYes{}
    \item[] Justification: The paper reports the computational resources used for model inference, video replay, and LLM-based evaluation, including GPU type, memory, runtime, and overall compute estimates. Additional resource details are provided in the appendix.
    \item[] Guidelines:
    \begin{itemize}
        \item The answer \answerNA{} means that the paper does not include experiments.
        \item The paper should indicate the type of compute workers CPU or GPU, internal cluster, or cloud provider, including relevant memory and storage.
        \item The paper should provide the amount of compute required for each of the individual experimental runs as well as estimate the total compute. 
        \item The paper should disclose whether the full research project required more compute than the experiments reported in the paper (e.g., preliminary or failed experiments that didn't make it into the paper). 
    \end{itemize}
    
\item {\bf Code of ethics}
    \item[] Question: Does the research conducted in the paper conform, in every respect, with the NeurIPS Code of Ethics \url{https://neurips.cc/public/EthicsGuidelines}?
    \item[] Answer: \answerYes{}
    \item[] Justification: The research follows the NeurIPS Code of Ethics by using publicly available or properly credited assets, filtering unsafe or private content, documenting annotation procedures, and discussing potential dual-use risks. The released assets are intended for research and evaluation purposes.
    \item[] Guidelines:
    \begin{itemize}
        \item The answer \answerNA{} means that the authors have not reviewed the NeurIPS Code of Ethics.
        \item If the authors answer \answerNo, they should explain the special circumstances that require a deviation from the Code of Ethics.
        \item The authors should make sure to preserve anonymity (e.g., if there is a special consideration due to laws or regulations in their jurisdiction).
    \end{itemize}

\item {\bf Broader impacts}
    \item[] Question: Does the paper discuss both potential positive societal impacts and negative societal impacts of the work performed?
    \item[] Answer: \answerYes{}
    \item[] Justification: The paper discusses positive impacts such as evaluating generative controllability, supporting creative reuse, and studying prompt leakage risks. It also discusses negative impacts such as potential prompt stealing, misuse for imitation, and risks related to intellectual property or privacy.
    \item[] Guidelines:
    \begin{itemize}
        \item The answer \answerNA{} means that there is no societal impact of the work performed.
        \item If the authors answer \answerNA{} or \answerNo, they should explain why their work has no societal impact or why the paper does not address societal impact.
        \item Examples of negative societal impacts include potential malicious or unintended uses (e.g., disinformation, generating fake profiles, surveillance), fairness considerations (e.g., deployment of technologies that could make decisions that unfairly impact specific groups), privacy considerations, and security considerations.
        \item The conference expects that many papers will be foundational research and not tied to particular applications, let alone deployments. However, if there is a direct path to any negative applications, the authors should point it out. For example, it is legitimate to point out that an improvement in the quality of generative models could be used to generate Deepfakes for disinformation. On the other hand, it is not needed to point out that a generic algorithm for optimizing neural networks could enable people to train models that generate Deepfakes faster.
        \item The authors should consider possible harms that could arise when the technology is being used as intended and functioning correctly, harms that could arise when the technology is being used as intended but gives incorrect results, and harms following from (intentional or unintentional) misuse of the technology.
        \item If there are negative societal impacts, the authors could also discuss possible mitigation strategies (e.g., gated release of models, providing defenses in addition to attacks, mechanisms for monitoring misuse, mechanisms to monitor how a system learns from feedback over time, improving the efficiency and accessibility of ML).
    \end{itemize}
    
\item {\bf Safeguards}
    \item[] Question: Does the paper describe safeguards that have been put in place for responsible release of data or models that have a high risk for misuse (e.g., pre-trained language models, image generators, or scraped datasets)?
    \item[] Answer: \answerYes{}
    \item[] Justification: The paper describes safeguards for responsible release, including filtering unsafe or private content, releasing the benchmark for research evaluation rather than misuse, and avoiding the release of tools designed to attack proprietary systems. The paper also discusses prompt leakage risks and mitigation considerations.
    \item[] Guidelines:
    \begin{itemize}
        \item The answer \answerNA{} means that the paper poses no such risks.
        \item Released models that have a high risk for misuse or dual-use should be released with necessary safeguards to allow for controlled use of the model, for example by requiring that users adhere to usage guidelines or restrictions to access the model or implementing safety filters. 
        \item Datasets that have been scraped from the Internet could pose safety risks. The authors should describe how they avoided releasing unsafe images.
        \item We recognize that providing effective safeguards is challenging, and many papers do not require this, but we encourage authors to take this into account and make a best faith effort.
    \end{itemize}

\item {\bf Licenses for existing assets}
    \item[] Question: Are the creators or original owners of assets (e.g., code, data, models), used in the paper, properly credited and are the license and terms of use explicitly mentioned and properly respected?
    \item[] Answer: \answerYes{}
    \item[] Justification: The paper credits the original creators of all existing datasets, models, APIs, and codebases used in the work. The appendix provides an asset table listing their sources, versions, citations, licenses, and terms of use when available.
    \item[] Guidelines:
    \begin{itemize}
        \item The answer \answerNA{} means that the paper does not use existing assets.
        \item The authors should cite the original paper that produced the code package or dataset.
        \item The authors should state which version of the asset is used and, if possible, include a URL.
        \item The name of the license (e.g., CC-BY 4.0) should be included for each asset.
        \item For scraped data from a particular source (e.g., website), the copyright and terms of service of that source should be provided.
        \item If assets are released, the license, copyright information, and terms of use in the package should be provided. For popular datasets, \url{paperswithcode.com/datasets} has curated licenses for some datasets. Their licensing guide can help determine the license of a dataset.
        \item For existing datasets that are re-packaged, both the original license and the license of the derived asset (if it has changed) should be provided.
        \item If this information is not available online, the authors are encouraged to reach out to the asset's creators.
    \end{itemize}

\item {\bf New assets}
    \item[] Question: Are new assets introduced in the paper well documented and is the documentation provided alongside the assets?
    \item[] Answer: \answerYes{}
    \item[] Justification: The paper introduces VI-Bench as a new benchmark and provides documentation for its data format, construction process, difficulty levels, evaluation protocol, annotation guideline, and intended use. The released assets are accompanied by instructions and metadata.
    \item[] Guidelines:
    \begin{itemize}
        \item The answer \answerNA{} means that the paper does not release new assets.
        \item Researchers should communicate the details of the dataset\slash code\slash model as part of their submissions via structured templates. This includes details about training, license, limitations, etc. 
        \item The paper should discuss whether and how consent was obtained from people whose asset is used.
        \item At submission time, remember to anonymize your assets (if applicable). You can either create an anonymized URL or include an anonymized zip file.
    \end{itemize}

\item {\bf Crowdsourcing and research with human subjects}
    \item[] Question: For crowdsourcing experiments and research with human subjects, does the paper include the full text of instructions given to participants and screenshots, if applicable, as well as details about compensation (if any)? 
    \item[] Answer: \answerYes{}
    \item[] Justification: The paper includes human validation with annotators and provides the full annotation instructions, scoring criteria, task examples, and compensation details in the appendix. The annotation protocol covers both prompt-level and video-level comparison tasks.
    \item[] Guidelines:
    \begin{itemize}
        \item The answer \answerNA{} means that the paper does not involve crowdsourcing nor research with human subjects.
        \item Including this information in the supplemental material is fine, but if the main contribution of the paper involves human subjects, then as much detail as possible should be included in the main paper. 
        \item According to the NeurIPS Code of Ethics, workers involved in data collection, curation, or other labor should be paid at least the minimum wage in the country of the data collector. 
    \end{itemize}

\item {\bf Institutional review board (IRB) approvals or equivalent for research with human subjects}
    \item[] Question: Does the paper describe potential risks incurred by study participants, whether such risks were disclosed to the subjects, and whether Institutional Review Board (IRB) approvals (or an equivalent approval/review based on the requirements of your country or institution) were obtained?
    \item[] Answer: \answerNA{}
    \item[] Justification: The annotation task only asks annotators to rate generated videos and prompts, does not collect personal or sensitive information, and does not study the annotators themselves. Therefore, IRB approval or equivalent review is not applicable under this setting.
    \item[] Guidelines:
    \begin{itemize}
        \item The answer \answerNA{} means that the paper does not involve crowdsourcing nor research with human subjects.
        \item Depending on the country in which research is conducted, IRB approval (or equivalent) may be required for any human subjects research. If you obtained IRB approval, you should clearly state this in the paper. 
        \item We recognize that the procedures for this may vary significantly between institutions and locations, and we expect authors to adhere to the NeurIPS Code of Ethics and the guidelines for their institution. 
        \item For initial submissions, do not include any information that would break anonymity (if applicable), such as the institution conducting the review.
    \end{itemize}

\item {\bf Declaration of LLM usage}
    \item[] Question: Does the paper describe the usage of LLMs if it is an important, original, or non-standard component of the core methods in this research? Note that if the LLM is used only for writing, editing, or formatting purposes and does \emph{not} impact the core methodology, scientific rigor, or originality of the research, declaration is not required.
    \item[] Answer: \answerYes{}
    \item[] Justification: The paper uses LLMs and VLMs as core components of the evaluation pipeline, including prompt-level judging and video-level memory-and-judge evaluation. The paper describes the model roles, input-output format, scoring protocol, and validation against human annotations.
    \item[] Guidelines:
    \begin{itemize}
        \item The answer \answerNA{} means that the core method development in this research does not involve LLMs as any important, original, or non-standard components.
        \item Please refer to our LLM policy in the NeurIPS handbook for what should or should not be described.
    \end{itemize}

\end{enumerate}

\end{document}